\documentclass[]{onethree}

\usepackage[toc,page,header]{appendix}
\usepackage{minitoc}
\usepackage{amsmath,amssymb}
\usepackage{bm}
\usepackage{wrapfig}
\usepackage{enumitem}
\usepackage{graphicx}
\usepackage{capt-of}
\usepackage{wrapfig}
\usepackage{needspace}
\usepackage{multirow}
\usepackage{booktabs}
\usepackage{enumitem}

\usepackage{amsmath,amsfonts,bm}

\def\eqref#1{equation~\ref{#1}}

\def\1{\bm{1}}

\DeclareMathAlphabet{\mathsfit}{\encodingdefault}{\sfdefault}{m}{sl}
\SetMathAlphabet{\mathsfit}{bold}{\encodingdefault}{\sfdefault}{bx}{n}

\usepackage{url}
\usepackage[table]{xcolor}
\usepackage{hyperref}
\usepackage{url}

\title{HoloRecon: Incremental Reconstruction of Long-Range Panoramas with Global Consistency}

\author[1,*]{Zhuoxiao Li}
\author[1,*]{Xinyi Liu}
\author[1,*]{Hongye Hou}
\author[1,*]{Jing OU}
\author[1]{Kang Du}
\author[1]{Shuai Zhang}
\author[1]{Dongli Wu}
\author[2]{Taoyu Wu}
\author[3]{Siyuan Feng}
\author[4]{Xiaohui Zhu}
\author[1,\dagger]{Wufan Zhao}

\affiliation[1]{The Hong Kong University of Science and Technology (GuangZhou)}
\affiliation[2]{University of Liverpool}
\affiliation[3]{Hong Kong Polytechnic University}
\affiliation[4]{Xi'an Jiaotong-Liverpool University}

\contribution[*]{Equal contribution.}
\contribution[\dagger]{Corresponding author.}

\abstract{
Long-sequence panoramic reconstruction requires detailed scene geometry and globally consistent camera poses across distant revisits. Panoramic SLAM and incremental SfM rely on feature correspondences, but repeated, similar structures and small partial overlap make reliable long-range connections difficult to establish, leaving accumulated drift unresolved. Feed-forward models offer detailed local geometry, but their high computational costs limit inference window size, necessitating additional constraints for global consistency. This motivates combining continuous feature-based tracking and sparse mapping with rapid, high-quality local reconstruction from feed-forward models to achieve accurate localization and globally consistent dense mapping over long sequences. We therefore propose a panoramic reconstruction framework that couples an ORB-based sparse panoramic mapping frontend with a fine-tuned feed-forward model. Our key insight is that predicted local geometry enables 3D verification of feature matches across distant revisits, yielding long-range camera constraints even between regions. We calibrate predicted depths with sparse landmarks and use verified connections to optimize the global pose graph, updating dense geometry with the corrected camera poses. We also introduce PanoRecon, a seven-scene benchmark with repetitive architecture, large and small loops, and reference camera poses and LiDAR geometry for evaluating global consistency. Our method achieves the lowest mean absolute trajectory error and high F-1 Score among the evaluated methods. Project Page: \href{https://zhuoxiaoli.github.io/HoloRecon_page/}{HoloRecon}. 
}

\teaser{
    \centering
    \includegraphics[width=1\linewidth]{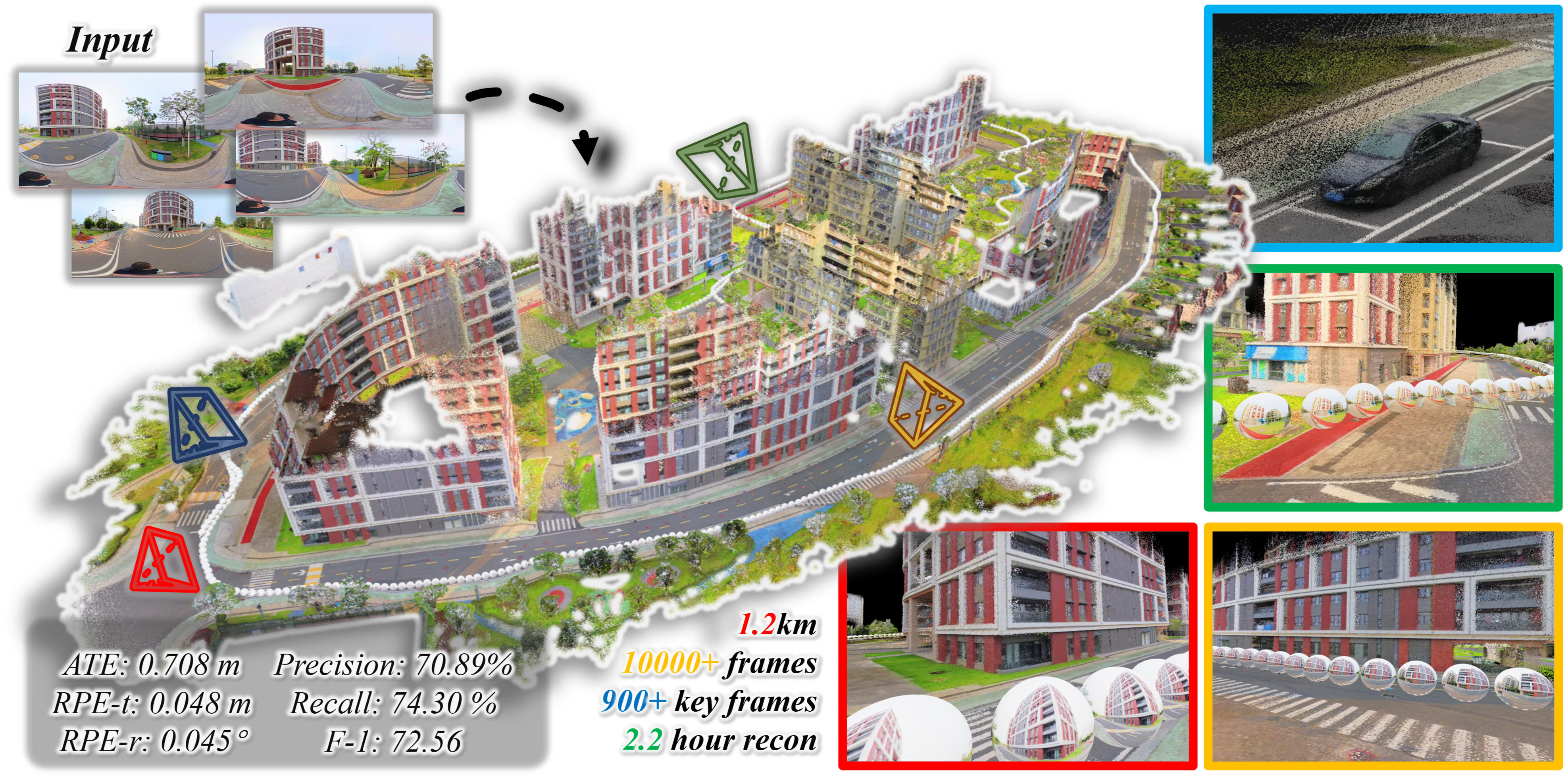}
    \caption{\textbf{Globally consistent reconstruction from a long panoramic sequence.} Our method reconstructs long-range panoramic sequence along a 1.2\,km route by coupling sparse mapping with feed-forward geometry.}
    \label{fig:placeholder}
}

\begin{document}
\maketitle



\section{Introduction}

Panoramic reconstruction recovers camera trajectories and dense geometry from equirectangular projection (ERP) images covering the full surroundings. Along extended routes through buildings and streets, reconstruction requires detailed surfaces and consistent placement of observations, particularly across distant revisits.

Feature-based panoramic SLAM and incremental structure-from-motion (SfM) register cameras and maintain sparse landmarks through image correspondences. SLAM systems use loop closure to correct accumulated drift, with acceptance thresholds controlling which revisits enter global optimization~\citep{surmann2022panorama,campos2021orbslam3}. Relaxed thresholds admit false associations, whereas strict thresholds reject valid loops with limited matching support. Repeated facade textures and symmetric layouts make distinct places visually similar, while distant revisits often share only partial overlap. False loops distort the map, and missed loops leave drift uncorrected, producing misaligned structural layers. Reliable global correction of panoramic reconstruction therefore requires stronger evidence for distinguishing true revisits from repeated scene patterns.

Feed-forward reconstruction rapidly estimates camera poses and detailed local geometry from panoramic views, as demonstrated by PanoVGGT~\citep{guo2026panovggtfeedforward3dreconstruction} and panoramic adaptations of Pi3~\citep{wang2026pi3permutationequivariantvisualgeometry}. Memory and computational costs limit the number of jointly processed views, while separate inference windows still require global constraints to remain consistent. {\textit{This motivates combining continuous feature-based tracking and mapping with rapid local feed-forward reconstruction for accurate localization and globally consistent  reconstruction over long sequences.}}

Based on the above mentioned discussion, we therefore introduce an incremental panoramic reconstruction framework that couples ORB-based sparse mapping with a fine-tuned feed-forward panoramic model (Fig.~\ref{fig:placeholder}). Our key insight is that \textbf{the predicted local geometry can enable 3D verification of feature matches across distant revisits}. We first calibrate predicted geometry with sparse landmarks and verify a common relative transform across views to establish long-range camera constraints. Accepted connections further drive pose-graph correction, which establishes shared observations for panoramic bundle adjustment with a sparse differentiation GPU backend. After refinement, we recalibrate cached depths and update each dense point using its source camera's corrected pose, without repeating inference.

To evaluate global consistency under these conditions, we further introduce the PanoRecon benchmark, comprising seven real-world scenes with repeated architectural textures and geometric structures. Every scene contains multiple revisits separated by long travel distances. We also collect LiDAR data, providing reference camera poses and dense geometry for evaluating trajectory accuracy and reconstruction consistency across distant revisits.

Our contributions are as follows:
\begin{itemize}[leftmargin=1.2em,itemsep=1pt,topsep=2pt,parsep=0pt]
    \item We propose  \textbf{a long-range panoramic reconstruction framework} that couples feature-based sparse mapping with feed-forward geometry to recover globally consistent camera trajectories and detailed dense maps.
    \item We introduce \textbf{a long-range global connection mechanism} that integrates feed-forward panoramic geometry with feature matching to enable 3D verification and establish camera constraints without requiring shared sparse landmarks between revisited regions.
    \item We introduce the \textbf{PanoRecon benchmark} with repetitive architecture and long-range revisits to evaluate trajectory accuracy and dense reconstruction consistency.
\end{itemize}

\section{Related Work}

\subsection{Structure from Motion}
SfM recovers cameras and structure through incremental registration~\citep{Schonberger_2016_CVPR,Heinly_2015_CVPR} or global motion estimation~\citep{Moulon_2013_ICCV,wilson_eccv2014_1dsfm,pan2024globalstructurefrommotionrevisited}. Resent study extends these operations to spherical imagery~\citep{dasilveira20243dscenegeometryestimation}, including spherical pose estimation and bundle adjustment~\citep{jiang20233dreconstructionsphericalimages}, dense matching~\citep{Jung_2025_ICCV}, and structural constraints from lines and architectural layouts~\citep{ikehata2016panoramicstructuremotiongeometric,huang2026savmapstructureaidedvisualmapping}. These advances improve registration and geometry estimation. Distant regions with partial overlap still require reliable associations for global refinement, motivating our use of predicted depths to support feature correspondences across views.

\subsection{Visual Odometry and Panoramic SLAM}
Visual odometry estimates sequential motion, with direct methods optimizing photometric error~\citep{engel2016directsparseodometry}, while feature-based, direct, and learned SLAM maintain maps with multi-view constraints~\citep{murartal2017orbslam2opensourceslammonocular,gao2018ldsodirectsparseodometry,teed2022droidslamdeepvisualslam}. OpenVSLAM~\citep{sumikura2023openvslamversatilevisualslam} supports purely visual equirectangular mapping. Visual--inertial estimation~\citep{qin2017vinsmonorobustversatilemonocular} extends to panoramic and spherical multi-camera observations~\citep{wang2022lfviovisualinertialodometryframeworklarge,wang2023lfvislamslamframeworklarge,wu2026panoairpanoramicvisualinertialslam,yang2026sphereviofastrobustvisualinertial}, while HDPV-SLAM~\citep{ahmadi2023hdpvslamhybriddepthaugmentedpanoramic} incorporates LiDAR depth. Their global accuracy depends on distinguishing true revisits from repeated appearance. 

\subsection{Feed-Forward Reconstruction}
Feed-forward geometry and matching models~\citep{wang2024dust3rgeometric3dvision,leroy2024groundingimagematching3d,wang2025vggtvisualgeometrygrounded,wang2026pi3permutationequivariantvisualgeometry} supply dense reconstruction priors. CUT3R maintains persistent state~\citep{wang2025continuous3dperceptionmodel}, SLAM3R learns local-to-world registration~\citep{liu2025slam3rrealtimedensescene}, and StreamVGGT and InfiniteVGGT use causal memory~\citep{zhuo2026streaming4dvisualgeometry,yuan2026infinitevggtvisualgeometrygrounded}. MASt3R-SLAM, VGGT-SLAM, VGGT-Long, and VGGT-SLAM 2.0 couple predictions with graph optimization and loop closure~\citep{murai2025mast3rslamrealtimedenseslam,maggio2025vggtslamdensergbslam,deng2026vggtlongchunkitloop,maggio2026vggtslam20realtimedense}. VGGT-Long processes overlapping chunks, while VGGT-SLAM 2.0 uses attention to verify revisits. For panoramas, PanoVGGT~\citep{guo2026panovggtfeedforward3dreconstruction} predicts poses and geometry, while HALO-SLAM~\citep{xiong2026looklookbackhidden} exploits attention, and dense geometry for tracking and loop verification.

\section{Method}
\label{sec:method}

\begin{figure}[!ht]
    \centering
    \includegraphics[width=\linewidth]{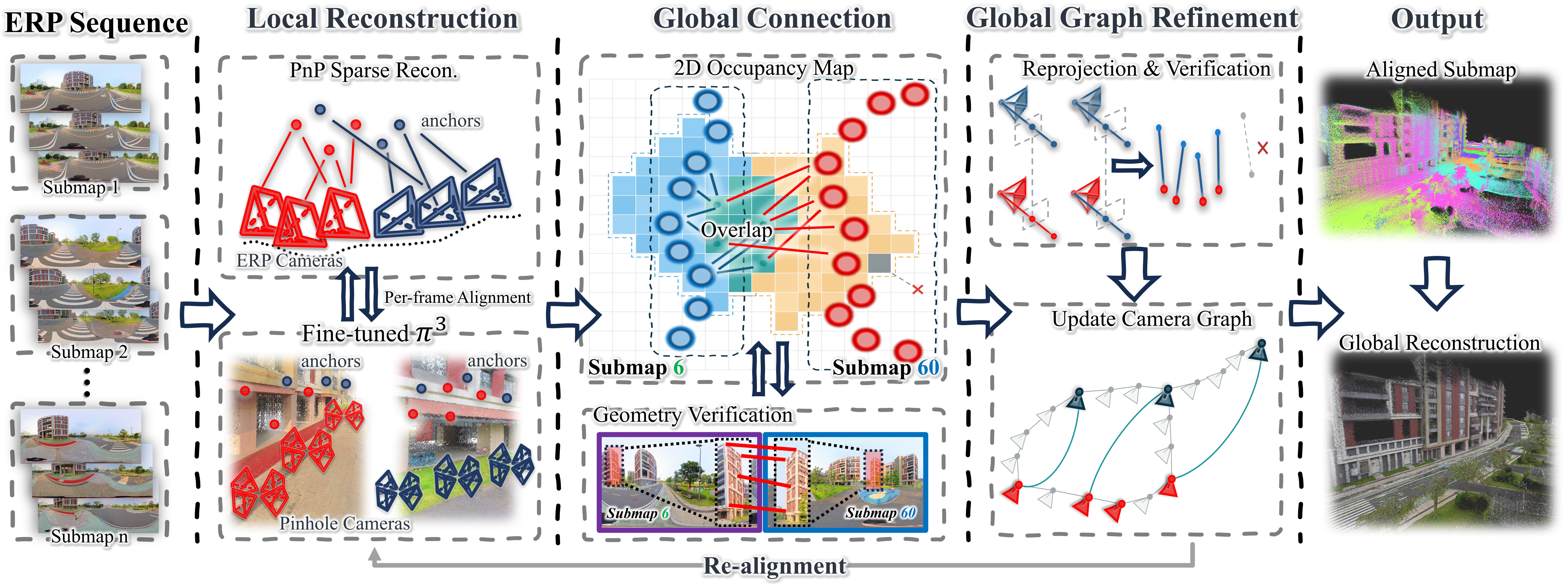}
    \vspace{-20pt}
    \caption{\textbf{Overview of the proposed method.} Sparse landmarks calibrate Pi3 depths, and registered panoramic cameras place them in the map. These predictions support cross-view ORB correspondences for verifying long-range camera constraints. Global correction and observation fusion refine the sparse map, which recalibrates and repositions the cached dense geometry.}
    \label{fig:method_pipeline}
\end{figure}

Given an ERP image sequence, we reconstruct camera poses and dense geometry by combining a ORB-based sparse map with local feed-forward predictions (Fig.~\ref{fig:method_pipeline}). We first calibrates and places the dense predictions in a common coordinate (Sec.~\ref{sec:method:calibration}). We then use these geometries to verify feature matches across views and construct long-range camera constraints (Sec.~\ref{sec:method:connections}). Optimizing the sparse map with these constraints updates the cameras and landmarks, which we use to recalibrate and reposition the cached dense points (Sec.~\ref{sec:method:refinement}).

\subsection{Anchoring Feed-Forward Geometry to the Sparse Map}
\label{sec:method:calibration}

\paragraph{Local predictions and the sparse map.}
An ORB-based ERP frontend \cite{surmann2022panorama} incrementally estimates camera-to-world poses $T_i=(R_i,\mathbf c_i)$ and sparse landmarks $\mathbf L_j$, where $i$ indexes keyframes, $j$ indexes landmarks, and $T_i\mathbf x=R_i\mathbf x+\mathbf c_i$. We group successive keyframes into overlapping windows and render their panoramas into calibrated perspective views. Pi3~\citep{wang2026pi3permutationequivariantvisualgeometry}, fine-tuned on Holo360D~\citep{ou2026holo360d}, jointly processes these views to predict local geometry. Each window defines a submap, whose predicted depths we anchor to the frontend's cameras and landmarks.

We additionally replace the per-view sky segmentation network with a lightweight sky head supervised by Holo360D geometry ground truth. It predicts a sky mask directly on each ERP image and projects to the perspective views, reducing inference time per submap from 40\,s to 13\,s.
\paragraph{Camera-anchored dense representation.}
Global optimization of ORB frontend can update the cameras within a submap differently. We therefore retain each prediction in its source camera coordinates and determine its world position from the current sparse-map pose. For pixel $\mathbf p$ in virtual view $v$ of keyframe $i$ and window $w$, let $\hat z_{wiv}(\mathbf p)$ be the optical-axis depth of the Pi3 local point. Its corresponding world point is
\begin{equation}
    \mathbf X_{wiv}(\mathbf p)
    =T_i\!\left(s_{wi}Q_v\hat z_{wiv}(\mathbf p)
                      K_v^{-1}\bar{\mathbf p}\right),
    \qquad s_{wi}=\exp(a_w+b_{wi}),
    \label{eq:method:dense-placement}
\end{equation}
where $\bar{\mathbf p}$ is the homogeneous pixel, $K_v$ is the rendering intrinsic matrix, and $Q_v$ rotates the virtual view into the panoramic camera. The virtual views share the panorama's optical center and fixed relative rotations. The shared window log scale $a_w$ aligns predicted depths to the sparse map, while the frame log-scale correction $b_{wi}$ captures variation within the window. The calibrated rays and sparse-map poses determine where points are placed, while depth calibration estimates their scales.

\paragraph{Submap alignment from landmarks.}
We estimate these scales from reliable landmark observations pooled within each window. For landmark $j$, its observation maps to pixel $\mathbf p_{ijv}$ in virtual view $v$, and the current pose gives its optical-axis depth $z^{\mathrm{geom}}_{ijv}$. Comparing this depth with the prediction gives the log ratio $y_{wijv}=\log[z^{\mathrm{geom}}_{ijv}/\hat z_{wiv}(\mathbf p_{ijv})]$. We fit the shared scale and the vector of frame scale corrections $\mathbf b_w=(b_{wi})_i$ through
\begin{equation}
    \min_{a_w,\mathbf b_w}\;
    \sum_{(i,j,v)\in\mathcal A_w}\rho(y_{wijv}-a_w-b_{wi})
    +\frac{\lambda_p}{2}\sum_i b_{wi}^{2}
    +\frac{\lambda_s}{2}\sum_{(i,k)\in\mathcal N_w}(b_{wi}-b_{wk})^{2}.
    \label{eq:method:depth-calibration}
\end{equation}
Here $\mathcal A_w$ contains reliable observations with positive geometric and predicted depths, $\rho$ is the Huber penalty, and $\mathcal N_w$ links adjacent frames with enough valid landmark observations. The weights $\lambda_p$ and $\lambda_s$ respectively constrain frame scale corrections toward the shared scale and encourage temporal smoothness. Log ratios measure proportional depth errors across scene distances. All frames share the window scale, with frame-specific corrections estimated from sufficient landmark support. After robust fitting, we align depths at common source pixels in overlapping windows, as detailed in Appendix~\ref{app:local-reconstruction}. Substituting the calibrated scales into Eq.~(\ref{eq:method:dense-placement}) expresses each submap in the ORB landmarks' coordinates for the connection estimation below.

\subsection{Verifying Long-Range Connections with Predicted Geometry}
\label{sec:method:connections}
\begin{wrapfigure}{r}{0.4\textwidth}
    \centering
    \includegraphics[width=\linewidth]{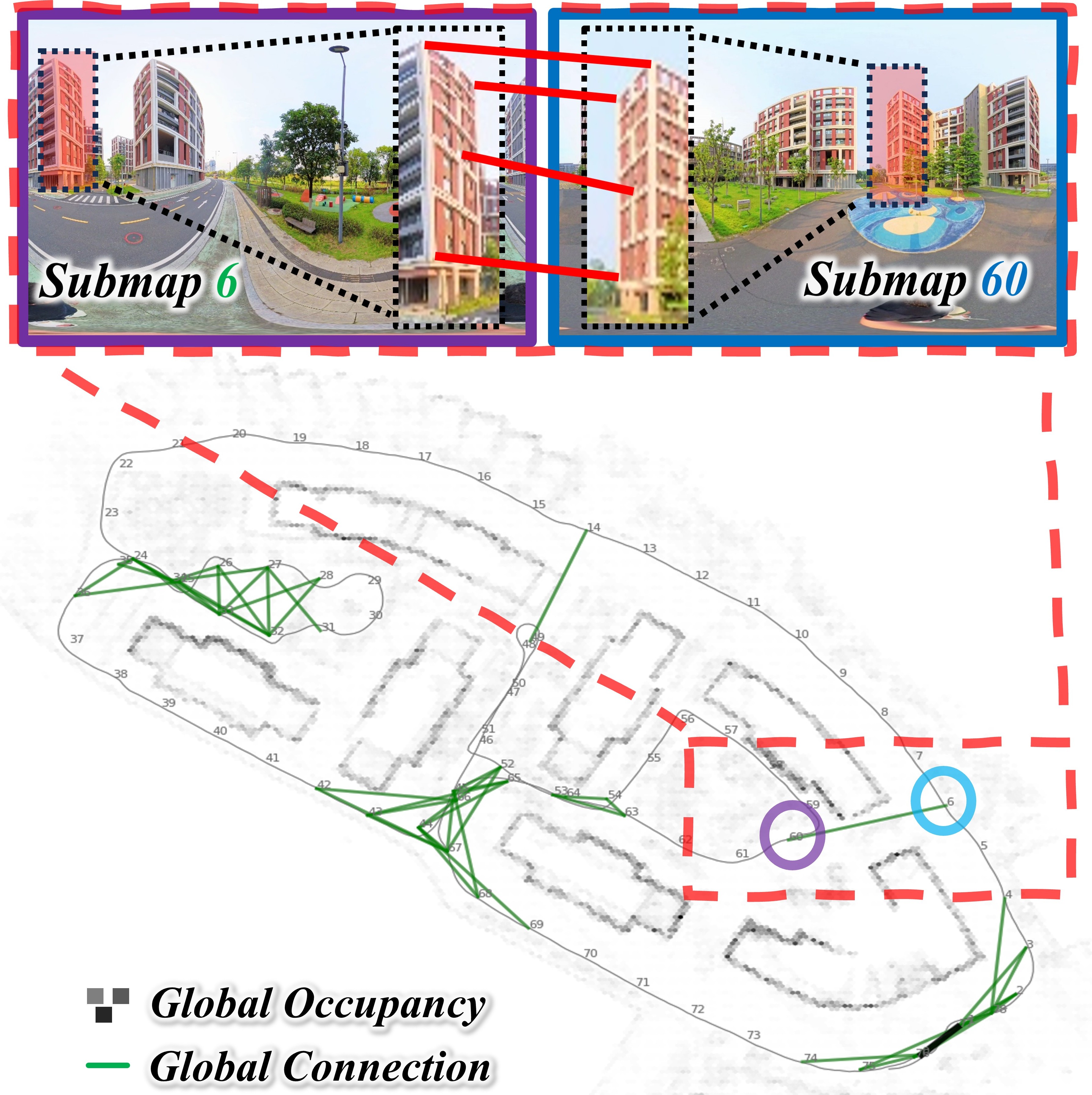}
    \vspace{-20pt}
    \caption{Cross-view evidence for a global connection.}
        \vspace{-20pt}
    \label{fig:method_connections}
\end{wrapfigure}

We use the aligned submaps to identify and verify overlap between distant observations. For a new submap $B$, we propose historical submaps $A$ using voxel overlap and line-of-sight visibility at their current map positions (Figure \ref{fig:method_connections}). Expanded spatial bounds and neighboring historical windows broaden this candidate set. Within each candidate pair, we compute mutual ORB matches across all pairs of keyframes with predicted depths. Combining matches across image pairs lets us use overlap distributed across views.

\paragraph{Depth-supported similarity estimation.}
To check their geometric consistency, we back-project matched keypoints using the calibrated depths from Sec.~\ref{sec:method:calibration} (Fig.~\ref{fig:method_connections}). A sample's distance from its source camera is $s_{wi}\hat z_{wiv}(\mathbf p)\|K_v^{-1}\bar{\mathbf p}\|_2$. Within each submap, we project samples into their source camera's ERP image and take the median distance near each feature, denoted $d_i(\mathbf u)$ for ERP pixel $\mathbf u$. With unit bearing $\mathbf b(\mathbf u)$, the matched feature becomes $\mathbf P_i(\mathbf u)=T_i[d_i(\mathbf u)\mathbf b(\mathbf u)]$ in world coordinates. The resulting point correspondences from keyframes $a\in A$ and $b\in B$ are combined to estimate $S_{A\leftarrow B}\in\operatorname{Sim}(3)$ by RANSAC and least-squares refitting~\citep{umeyama1991transformation}, and define submaps' rotation, translation, and positive scale. It aligns corresponding points through $\mathbf P_a\approx S_{A\leftarrow B}\mathbf P_b$, correcting the displacement caused by drift despite their shared map coordinates.

\paragraph{Cross-view geometric verification.}
Repeated structures can produce plausible local matches, so we test whether a common similarity explains correspondences across multiple views. We check bidirectional distance and bearing agreement, nondegenerate geometry, and distinct feature support from multiple keyframes on both sides. For large corrections, we divide the current submap's keyframes into groups with balanced numbers of matches before fitting. We estimate a transform from a group's matches and verify it using matches in the other, then repeat with the groups exchanged. Both transforms must agree with the estimate from all matches. Verified transforms from neighboring historical submaps must also agree before acceptance. Appendix~\ref{app:global-connections} specifies these checks and their thresholds.
\paragraph{Camera constraint construction.}
We convert each verified submap transform into relative camera constraints for the sparse map. We select image pairs with enough inliers and distinct camera endpoints, then form the relative constraint mapping camera $a$ to camera $b$,
\begin{equation}
    \widehat G_{b\leftarrow a}=T_b^{-1}S_{A\leftarrow B}^{-1}T_a,
    \label{eq:method:camera-measurement}
\end{equation}
using the camera poses at the time the constraint is formed. The measurement remains fixed during subsequent optimization. The inverse similarity maps points reconstructed from submap A onto those from submap B before conversion into camera $b$. We cap the total weight for each verified submap transform based on its inlier count and divide this weight among the selected camera constraints. These weighted constraints enter the pose graph below.

\subsection{Propagating Global Corrections to Sparse and Dense Geometry}
\label{sec:method:refinement}

\paragraph{Graph correction and observation fusion.}
We insert the relative camera constraints from Eq.~(\ref{eq:method:camera-measurement}) into the similarity pose graph alongside existing sparse-map constraints. We optimize the graph, remove incompatible new edges, and re-solve to update camera poses and landmark positions. In the corrected geometry, standard projection-based landmark fusion~\citep{murartal2017orbslam2opensourceslammonocular} runs bidirectionally between the cameras connected by accepted constraints and a limited number of their covisible neighbors. Merging duplicate landmarks establishes observations shared across revisited regions. We denote the updated camera--landmark observation set by $\mathcal O$, which supplies cross-region reprojection factors for global BA.

\paragraph{GPU-accelerated panoramic bundle adjustment.}
The shared observations allow BA to jointly refine the graph-corrected poses $T_i$ and landmarks $\mathbf L_j$. We use the original ERP feature pixels $(u_{ij},v_{ij})$, retaining one pose per panorama so that BA directly refines the sparse map used to align dense predictions. As long-range connections enlarge this observation graph, repeated global refinement becomes costly, motivating GPU acceleration. With predicted pixels $(\hat u_{ij},\hat v_{ij})=\pi_{\mathrm{ERP}}(T_i^{-1}\mathbf L_j)$, where $\pi_{\mathrm{ERP}}$ is the panoramic projection, we minimize
\begin{equation}
    \min_{\{T_i,\mathbf L_j\}} E_{\mathrm{BA}}
    =\sum_{(i,j)\in\mathcal O}\psi_{\delta_b}\!\left(\sigma_{ij}^{-2}\|\mathbf r_{ij}\|_2^2\right),
    \qquad \mathbf r_{ij}=\bigl[\operatorname{wrap}_W(u_{ij}-\hat u_{ij}),\;v_{ij}-\hat v_{ij}\bigr]^\top.
\label{eq:method:erp-ba}
\end{equation}
Here $\psi_{\delta_b}$ is the Huber loss on the squared weighted residual norm, with transition at $\delta_b^2$, and $\sigma_{ij}^{-2}$ is the feature-pyramid inverse variance. For ERP width $W$, $\operatorname{wrap}_W(q)=q-W\lfloor q/W+1/2\rfloor$ measures horizontal error across the image seam. Starting from the corrected geometry with the root pose fixed, sparse automatic differentiation~\citep{zhan2026bundle} linearizes the residuals. After landmark elimination, GPU preconditioned conjugate gradients solve the reduced camera system without storing the explicit Schur complement. Appendix~\ref{app:gpu-ba} details the solver, and Table~\ref{tab:ablation_ba_backend} reports its efficiency.

\paragraph{Updating cached dense geometry.}
The refined cameras and landmarks also change the depth references used in Sec.~\ref{sec:method:calibration}. We recompute the landmark depths, refit the scales in Eq.~(\ref{eq:method:depth-calibration}), and place the cached predictions using the updated $T_i$ and $s_{wi}$ in Eq.~(\ref{eq:method:dense-placement}). Each dense point is updated using its source camera's optimized pose and recalibrated depth, without repeating Pi3 inference. At each submap boundary, any triggered global refinement and dense updates finish before frame input resumes, so subsequent tracking and connection estimation use the corrected map.

\section{The PanoRecon Benchmark}
\label{sec:panorecon}

\begin{figure}[t]
    \centering
    \begin{minipage}[c]{0.5\linewidth}
        \centering
        \small
        \renewcommand{\arraystretch}{1.3}
        \setlength{\tabcolsep}{2pt}
        \begin{tabular}{@{}p{0.52\linewidth}p{\dimexpr0.48\linewidth-3pt\relax}@{}}
            \toprule
            \textbf{Dataset property} & \textbf{Value} \\
            \midrule
            Number of scenes & 7 \\
            Mean video duration & 11.49 min \\
            Frame sampling interval & 0.1 s \\
            Mean trajectory length (km) & 0.87 (0.26--1.69) \\
            Reprojection RMSE (pix) & 0.1917  \\
            Loop structure & Large and small loops \\
            \bottomrule
        \end{tabular}
        \par\smallskip {\small (a) Dataset statistics}
    \end{minipage}\hfill
    \begin{minipage}[c]{0.40\linewidth}
        \centering
        \includegraphics[width=\linewidth]{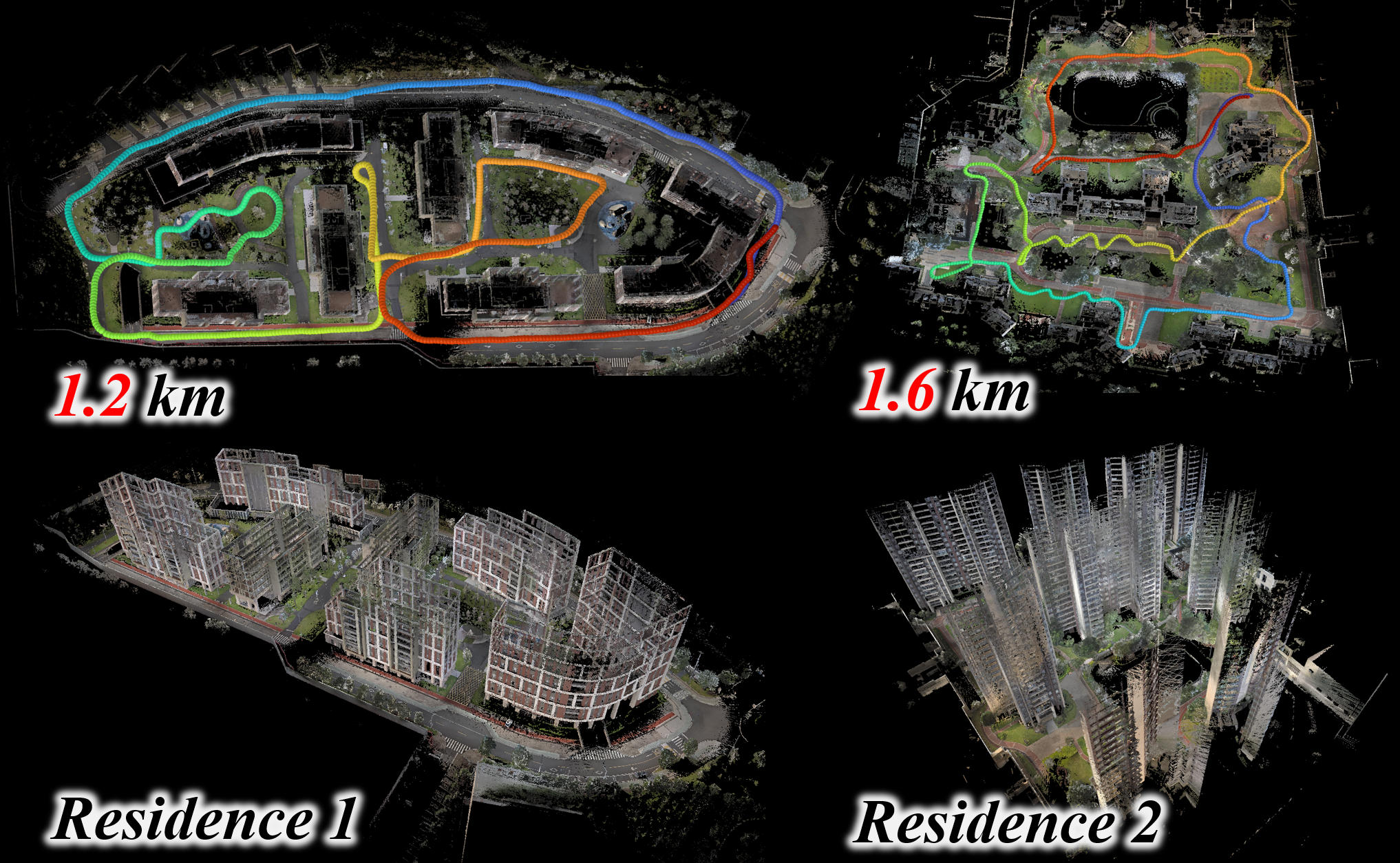}
        \par\smallskip {\small (b) Dataset example}
    \end{minipage}
    \vspace{-10pt}
    \caption{Dataset statistics and GT examples of the PanoRecon benchmark.}
    \label{fig:panorecon_benchmark}
\end{figure}
We independently capture panoramic videos and RTK-assisted LiDAR data along the same planned routes in seven scenes with low-texture surfaces and repeated architecture. Each route includes a large loop and distant revisits with limited overlap, testing reconstruction methods' ability to correct accumulated drift.

\paragraph{Reference geometry and camera poses.}\label{sec:panorecon:processing}
We optimize the LiDAR trajectory with RTK corrections and export the point cloud and calibrated poses of its three onboard cameras. In \textit{RealityScan}, these define LiDAR reference images with fixed poses. ERP frames sampled every 0.1\,s are split into calibrated perspective rigs and globally matched to these images. We optimize only the ERP rig poses, then recover one pose per panorama using the rig extrinsics. The reference poses share the LiDAR's metric coordinate system. The mean reprojection RMSE is 0.1917 pixels (Fig.~\ref{fig:panorecon_benchmark}). Appendix~\ref{app:panorecon} details acquisition and processing.

\paragraph{Reconstruction evaluation.}
For a reconstruction in frame $E$, trajectory alignment to the reference poses gives the similarity $S_{L\leftarrow E}$ into the LiDAR frame $L$. We apply it to the reconstructed geometry and refine registration with ICP. Following Tanks and Temples~\citep{Knapitsch_2017}, bidirectional nearest-neighbor distances yield precision, recall, and F-score. The complete transformation is
\begin{equation}
    \mathbf x_L^{\mathrm{eval}}
    = H_{\mathrm{ICP}}\,S_{L\leftarrow E}\,\mathbf x_E,
    \label{eq:panorecon:evaluation-alignment}
\end{equation}
where $H_{\mathrm{ICP}}$ refines cloud alignment for geometric evaluation.

\section{Experiments}
\paragraph{Implementation.}
We process submaps of 16 panoramic keyframes with four shared keyframes between consecutive submaps. Depth scale calibration uses a log-ratio Huber threshold of $0.05$. Connection verification uses bidirectional distance and bearing thresholds of $0.3$ map units and $1^\circ$, with support from at least three keyframes on each side. GPU bundle adjustment uses double precision and at most 250 PCG iterations with a tolerance of $10^{-5}$. Further settings are provided in Appendix~\ref{app:method-details}. All experiments are conducted on a single NVIDIA RTX PRO 6000 Blackwell Workstation Edition GPU.

\definecolor{groupgray}{RGB}{245,246,248} \definecolor{rankgreen}{HTML}{59A14F}
\newcommand{\first}[1]{\cellcolor{rankgreen!35}\textbf{#1}}
\newcommand{\second}[1]{\cellcolor{rankgreen!22}#1}
\newcommand{\third}[1]{\cellcolor{rankgreen!12}#1}
\begin{table}[!h]
\small
\setlength\tabcolsep{10pt}
\centering
\renewcommand\arraystretch{1.0}
\caption{\textbf{Quantitative comparison of localization.} Mean scene-wise RMSE on 360LOC (4 scenes), YUTO MMS (4 sequences), and PanoRecon (7 scenes). OOM denotes GPU out-of-memory.}
\label{tab:anchor_reconstruction}
\resizebox{\linewidth}{!}{
\begin{tabular}{l *{9}{c}}
\toprule
\multirow{2}{*}{Method} & \multicolumn{3}{c}{360LOC} & \multicolumn{3}{c}{YUTO MMS} & \multicolumn{3}{c}{PanoRecon} \\
\cmidrule(lr){2-4}\cmidrule(lr){5-7}\cmidrule(lr){8-10}
 & ATE$\downarrow$ & RPE-t$\downarrow$ & RPE-r$\downarrow$ & ATE$\downarrow$ & RPE-t$\downarrow$ & RPE-r$\downarrow$ & ATE$\downarrow$ & RPE-t$\downarrow$ & RPE-r$\downarrow$ \\
\midrule
\multicolumn{10}{>{\columncolor{groupgray}}l}{\textbf{Feed-Forward}} \\
PanoVGGT (ERP) & OOM & OOM & OOM & OOM & OOM & OOM & OOM & OOM & OOM \\
Scal3R (Pinhole) & 5.366 & 0.124 & 1.354 & 187.407 & 2.911 & 1.025 & 9.794 & 0.282 & 0.340 \\
\midrule
\multicolumn{10}{>{\columncolor{groupgray}}l}{\textbf{SLAM}} \\
360DVO (ERP) & 0.954 & 0.026 & 0.039 & \third{68.313} & \second{0.183} & \first{0.011} & \third{2.799} & \second{0.104} & \second{0.039} \\
Open-vSLAM (ERP) & 0.441 & 0.711 & 22.184 & 281.793 & 41.690 & 37.574 & 7.399 & 6.580 & 25.242 \\
Stella-vSLAM (ERP) & 0.706 & 0.047 & 0.147 & 82.173 & \third{0.656} & 0.131 & 7.424 & 1.329 & 0.470 \\
\midrule
\multicolumn{10}{>{\columncolor{groupgray}}l}{\textbf{Structure-from-Motion}} \\
COLMAP (ERP) & \second{0.355} & \first{0.020} & \second{0.037} & 124.490 & 0.814 & \third{0.066} & 3.510\,{\scriptsize (6/7)} & 2.575\,{\scriptsize (6/7)} & 1.452\,{\scriptsize (6/7)} \\
COLMAP (Pinhole) & \third{0.399} & \second{0.021} & \first{0.037} & 113.317 & 25.949 & 1.046 & \second{1.736} & \first{0.099} & \first{0.036} \\
Spherical-SfM (ERP) & 0.491 & \third{0.024} & \third{0.038} & \second{21.186} & 3.312 & 3.989 & 11.779 & 6.915 & 1.418 \\
\midrule
\textbf{Ours} (ERP) & \first{0.270} & 0.030 & 0.044 & \first{3.527} & \first{0.052} & \second{0.022} & \first{1.697} & \third{0.111} & \third{0.086} \\
\bottomrule
\end{tabular}}
\end{table}
\begin{figure}[!h]
    \centering
    \includegraphics[width=1\linewidth]{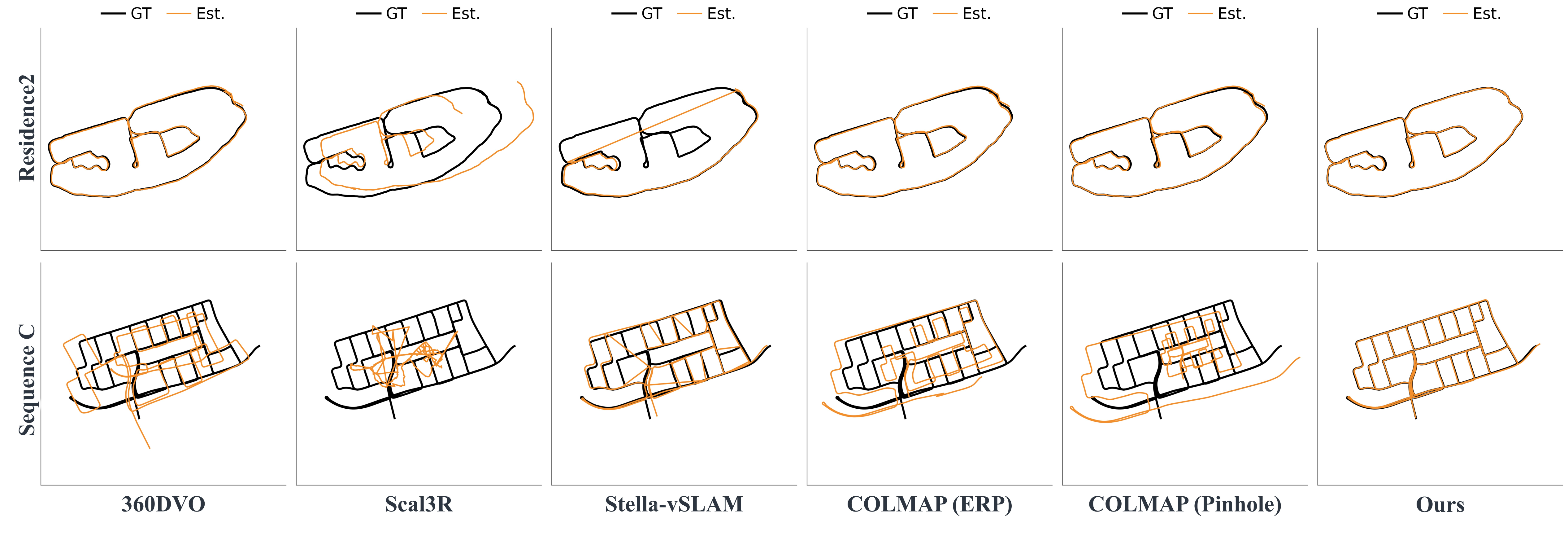}
    \vspace{-20pt}
    \caption{Qualitative comparison of localization.}
    \label{fig:traj_com}
\end{figure}

\paragraph{Datasets.}
\label{sec:experiments:datasets}
We use the mapping sequences from all four scenes of 360LOC~\citep{Huang_2024_CVPR}: Atrium, Concourse, Hall, and Piatrium. We also evaluate all four sequences (A--D) of YUTO MMS~\citep{Zhang_2024}, a panoramic dataset covering 20.1\,km  and the Sequence C alone spans 9.137\,km with more than 9,000 ERP images. Finally, we evaluate the seven PanoRecon scenes introduced in Sec.~\ref{sec:panorecon}. Both YUTO MMS and the PanoRecon benchmark contain long trajectories and complex loop structures, which challenge sustained tracking and global drift correction. 

\paragraph{Baselines.}
\label{sec:experiments:baselines}
We compare against PanoVGGT~\citep{guo2026panovggtfeedforward3dreconstruction} and Scal3R~\citep{Xie_2026_CVPR} for feed-forward reconstruction, 360DVO~\citep{11358682}, Open-vSLAM~\citep{sumikura2023openvslamversatilevisualslam}, and Stella-vSLAM-Dense \citep{surmann2022panorama} for panoramic visual odometry and SLAM, and COLMAP~\citep{Schonberger_2016_CVPR} and Spherical-SfM for SfM (Table~\ref{tab:anchor_reconstruction}). Scal3R receives distortion-free perspective views rendered from ERP images using COLMAP's official panorama renderer, with a $90^\circ$ field of view (front view). We evaluate COLMAP in two configurations: directly on ERP images with an equirectangular camera model, and on 12 such perspective overlapping views per panorama with known intrinsics and fixed per-ERP rig extrinsics.

\paragraph{Metrics.}
We evaluate camera poses after one global Umeyama $\mathrm{Sim}(3)$ alignment to the reference. ATE, translational RPE (RPE-t), and rotational RPE (RPE-r) are reported as RMSE. For dense reconstruction, we apply the trajectory similarity to the point cloud. We report mean scene-wise precision, recall, and F-score at 20\,cm on 360LOC, YUTO MMS, and PanoRecon (Table~\ref{tab:dense_reconstruction}). Appendices~\ref{app:localization-results} and~\ref{app:dense-results} provide per-scene results and evaluation details.

\begin{table}[!h]
\small
\setlength\tabcolsep{10pt}
\centering
\renewcommand\arraystretch{1.0}
\caption{\textbf{Quantitative comparison of dense reconstruction.} Mean scene-wise precision (P), recall (R), and F-score (F) in percent, at 20\,cm on 360LOC (4 scenes), YUTO MMS (4 sequences), and PanoRecon (7 scenes). Higher is better.}
\label{tab:dense_reconstruction}
\resizebox{\linewidth}{!}{
\begin{tabular}{l *{9}{c}}
\toprule
\multirow{2}{*}{Method} & \multicolumn{3}{c}{360LOC} & \multicolumn{3}{c}{YUTO MMS} & \multicolumn{3}{c}{PanoRecon} \\
\cmidrule(lr){2-4}\cmidrule(lr){5-7}\cmidrule(lr){8-10}
 & P$\uparrow$ & R$\uparrow$ & F$\uparrow$ & P$\uparrow$ & R$\uparrow$ & F$\uparrow$ & P$\uparrow$ & R$\uparrow$ & F$\uparrow$ \\
\midrule
Stella-vSLAM-Dense (ERP) & \third{62.42} & \first{58.23} & \second{58.68} & \third{23.78} & \second{14.52} & \third{17.24} & \third{36.61} & \third{31.10} & \third{31.97} \\
COLMAP (Pinhole) & \first{81.27} & \third{44.79} & \third{57.25} & \second{29.15} & \third{13.23} & \second{17.28} & \first{71.32} & \second{51.83} & \second{59.75} \\
\textbf{Ours} (ERP) & \second{79.22} & \second{52.87} & \first{62.04} & \first{60.56} & \first{40.77} & \first{48.71} & \second{64.22} & \first{62.36} & \first{63.16} \\
\bottomrule
\end{tabular}}
\end{table}

\begin{figure}[t]
    \centering
    \includegraphics[width=1\linewidth]{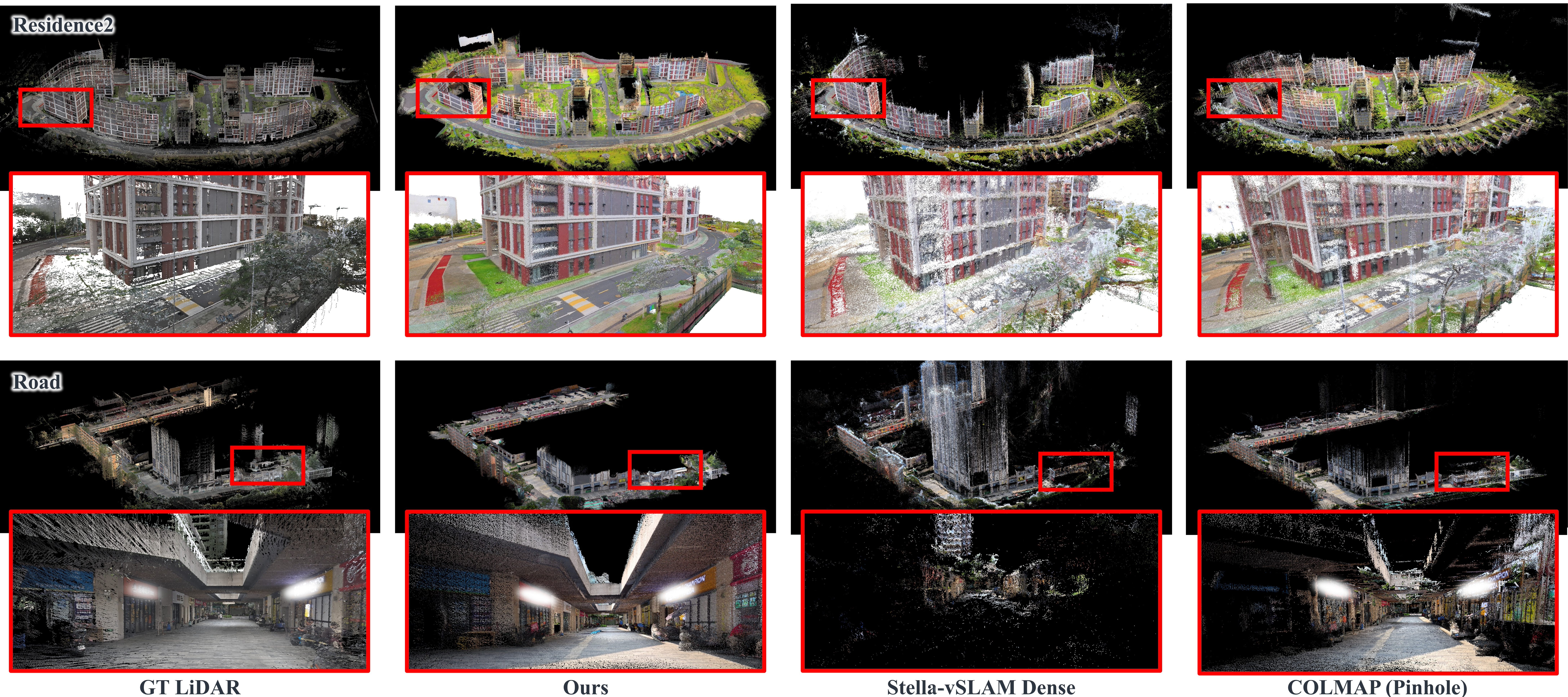}
    \vspace{-20pt}
    \caption{Point-cloud comparison on the Residence2 and Road scenes of the PanoRecon benchmark.}
    \label{fig:dense_comparison}
\end{figure}

\subsection{Trajectory Comparison}
\label{sec:experiments:trajectory}
Our method achieves the lowest mean ATE on all three datasets, with the largest advantage on YUTO MMS (Table~\ref{tab:anchor_reconstruction}). False loop closures distort ORB-based Stella-vSLAM trajectories on Residence2 and Sequence C, where buildings share similar appearance and geometry (Fig.~\ref{fig:traj_com}). Scal3R exhibits scale drift on large scenes with one perspective view per panorama, as does COLMAP during sequential reconstruction in both ERP and pinhole configurations. Our trajectories follow the reference street layouts and loop structures. On our single GPU, PanoVGGT~\citep{guo2026panovggtfeedforward3dreconstruction} ran out of memory (OOM) on every evaluated scene under the input settings in Appendix~\ref{app:baseline-settings}. HALO-SLAM~\citep{xiong2026looklookbackhidden} had no public implementation at evaluation time.

\subsection{Dense Reconstruction Comparison}
\label{sec:experiments:reconstruction}
Accurate camera poses keep local predictions aligned across distant revisits, reducing drift-induced misalignment in the global point cloud. Our largest F-score gains occur on the long-range YUTO MMS sequences (Table~\ref{tab:dense_reconstruction}), consistent with the trajectory results. Feed-forward geometry also improves local completeness on weakly textured walls and reflective surfaces, where photometric matching is unreliable. In these regions, the PatchMatch-based COLMAP and Stella-vSLAM-Dense baselines exhibit holes and noisy geometry, while our method recovers more continuous surfaces (Fig.~\ref{fig:dense_comparison} and supplementary Figs.~\ref{fig:dense_building}--\ref{fig:dense_hpc}). Appendix~\ref{app:dense-results} reports per-scene results.

\subsection{Runtime.}
\noindent
\begin{minipage}[t]{0.52\linewidth}
\vspace{0pt}
Appendix~\ref{app:runtime} provides the runtime analysis and stage-wise comparison on YUTO MMS Sequence C.
The GPU backend reduces the cost of global refinement while preserving comparable final reprojection error (Table~\ref{tab:ablation_ba_backend}). Matrix-free Schur optimization makes the growing observation graph more efficient to solve, supporting repeated correction as long-range connections accumulate.
\end{minipage}\hfill
\begin{minipage}[t]{0.40\linewidth}
\vspace{0pt}
\centering\scriptsize
\setlength{\tabcolsep}{2pt}
\setlength{\abovecaptionskip}{0pt}
\setlength{\belowcaptionskip}{4pt}
\captionof{table}{\textbf{CPU and GPU bundle adjustment.}}
\label{tab:ablation_ba_backend}
\resizebox{\linewidth}{!}{%
\begin{tabular}{@{}lrrrr@{}}
\toprule
Backend & \shortstack{Solve\\(s)$\downarrow$} & \shortstack{Total BA\\(s)$\downarrow$} & Speedup$\uparrow$ & \shortstack{Reproj.\\RMSE$\downarrow$} \\
\midrule
\shortstack[l]{CPU (OpenMP,\\32 threads)} & 19.430 & 32.649 & $1.00\times$ & 0.8466 \\
\shortstack[l]{GPU\\(CUDA Schur)} & \textbf{2.244} & \textbf{7.581} & $\mathbf{4.31\times}$ & \textbf{0.8420} \\
\bottomrule
\end{tabular}
}
\end{minipage}\par
\subsection{Ablation Studies}
\label{sec:ablations}
\paragraph{Effect of global constraints and bundle adjustment.}
In the Residence2 submap experiment, BA alone changes little, while a single camera constraint with BA provides most of the ATE reduction (Table~\ref{tab:ablation_global_connections}). Additional constraints improve ATE further. Graph correction improves global alignment, and BA restores local motion consistency, showing how a few constraints can benefit reconstruction through joint refinement.

\newsavebox{\staffAblationTable}
\newsavebox{\staffAblationCaption}
\begin{table}[t]
\centering
\begin{lrbox}{\staffAblationTable}
\begin{minipage}{0.5\linewidth}
\setlength{\abovecaptionskip}{0pt}
\setlength{\belowcaptionskip}{4pt}
\caption{\textbf{Effect of global constraints.} (Residence2 full trajectory)}
\label{tab:ablation_staff_backend}
\centering\scriptsize
\setlength{\tabcolsep}{1pt}
\resizebox{\linewidth}{!}{%
\begin{tabular}{@{}lrrrrr@{}}
\toprule
Configuration & \shortstack{ATE\\(m)$\downarrow$} & \shortstack{RPE-t\\(m)$\downarrow$} & \shortstack{RPE-r\\($^\circ$)$\downarrow$} & \shortstack{Reproj.\\(px)$\downarrow$} & \shortstack{Long-range\\observations} \\
\midrule
No connection & 1.358 & {0.061} & \textbf{0.045} & 1.623 & 160,186 \\
PG & 0.977 & 0.060 & 0.046 & 1.659 & 160,186 \\
PG + BA & 0.922 & 0.057 & \textbf{0.045} & \textbf{1.620} & 160,186 \\
PG + fusion & 0.877 & 0.060 & 0.046 & 1.760 & \textbf{293,827} \\
PG + fusion + BA & \textbf{0.708} & \textbf{0.048}  & \textbf{0.045} & 1.662 & \textbf{293,827} \\
\bottomrule
\end{tabular}
}
\end{minipage}
\end{lrbox}
\begin{lrbox}{\staffAblationCaption}
\begin{minipage}{0.5\linewidth}
\setlength{\abovecaptionskip}{4pt}
\setlength{\belowcaptionskip}{0pt}


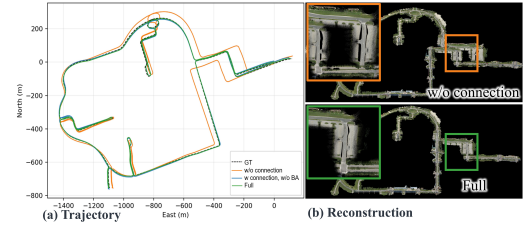
\captionof{figure}{Effect of global constraints.}
\label{fig:ablation_global_correction}
\end{minipage}
\end{lrbox}
\raisebox{-\height}{\usebox{\staffAblationTable}}\hfill
\raisebox{-\height}{%
\begin{minipage}[b][\dimexpr\ht\staffAblationTable+\dp\staffAblationTable\relax][s]{0.5\linewidth}
\centering
\includegraphics[width=\linewidth,height=\dimexpr\ht\staffAblationTable+\dp\staffAblationTable-\ht\staffAblationCaption-\dp\staffAblationCaption\relax,keepaspectratio]{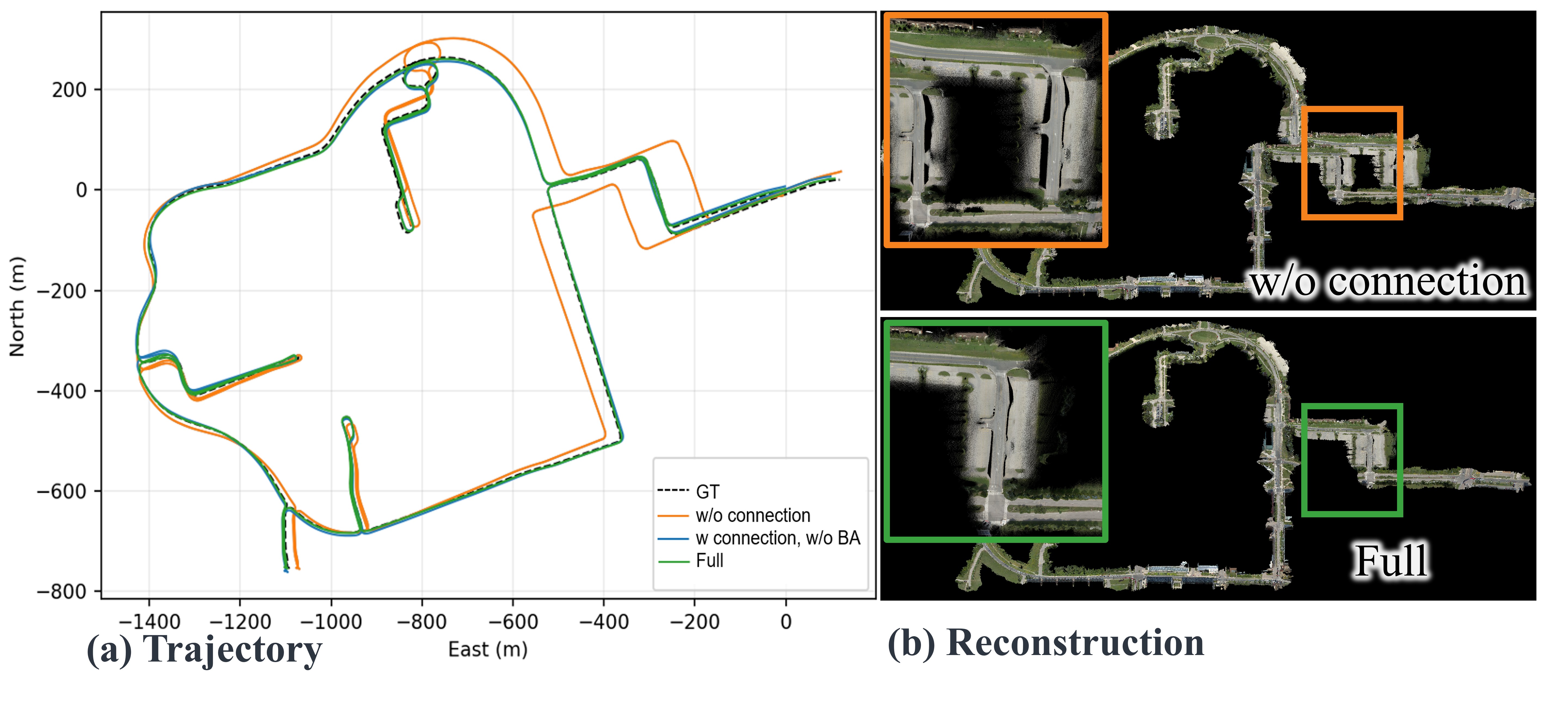}\par
\vfill
\nointerlineskip
\usebox{\staffAblationCaption}
\end{minipage}%
}
\end{table}

\begin{table}[t]
\centering
\setlength{\tabcolsep}{0.8pt}
\setlength{\abovecaptionskip}{0pt}
\setlength{\belowcaptionskip}{5pt}
\begin{minipage}[t]{0.345\linewidth}
\vspace{0pt}
\centering\footnotesize
\begin{minipage}[t][3\baselineskip][t]{\linewidth}
\caption{\textbf{Global connections and joint refinement.}( Submap $6\!\rightarrow\!60$; $16\!\rightarrow\!49$)}
\label{tab:ablation_global_connections}
\end{minipage}\par
\fontsize{6}{7}\selectfont
\renewcommand{\arraystretch}{1.32}
\begin{tabular*}{\linewidth}{@{\extracolsep{\fill}}lllrrrr@{}}
\toprule
Edges & Weights & BA & \shortstack{ATE\\(m)$\downarrow$} & \shortstack{RPE-t\\(m)$\downarrow$} & \shortstack{RPE-r\\($^\circ$)$\downarrow$} & \shortstack{F-1\\(\%)$\uparrow$} \\
\midrule
0 & -- & Off & 0.370 & 0.0128 & 0.0177 & 68.36 \\
0 & -- & On & 0.371 & 0.0129 & \textbf{0.0175} & 68.36 \\
5 & Inlier & Off & 0.296 & 0.0633 & 0.0939 & 71.76 \\
1 & Inlier & On & 0.185 & 0.0095 & 0.0176 & 71.49 \\
2 & Inlier & On & 0.180 & \textbf{0.0093} & 0.0176 & 71.59 \\
5 & Uniform & On & 0.170 &0.0094 & 0.0176 & 71.81 \\
5 & Inlier & On & \textbf{0.165} & 0.0094 & 0.0176 & \textbf{71.97} \\
\bottomrule
\end{tabular*}
\end{minipage}\hfill
\begin{minipage}[t]{0.28\linewidth}
\vspace{0pt}
\centering\footnotesize
\begin{minipage}[t][3\baselineskip][t]{\linewidth}
\caption{\textbf{Submap alignment.} (Submap $6\!\rightarrow\!60$; $16\!\rightarrow\!49$)}
\
\label{tab:ablation_local_anchoring}
\end{minipage}\par
\fontsize{6}{7}\selectfont
\renewcommand{\arraystretch}{1.87}
\begin{tabular*}{\linewidth}{@{\extracolsep{\fill}}p{0.49\linewidth}rrr@{}}
\toprule
\raggedright Alignment & P$\uparrow$ & R$\uparrow$  & $F-1\uparrow$ \\
\midrule
\raggedright Pose-derived scale & 85.14 & 61.22  & 71.23 \\
\raggedright Whole-submap Sim(3) & 85.08 & \textbf{61.93}  & 71.68 \\
\raggedright Sparse window scale & 86.16 & 60.72  & 71.24 \\
\raggedright Sparse per-frame scale (ours) & \textbf{86.48} & 61.62  & \textbf{71.97} \\
\bottomrule
\end{tabular*}
\end{minipage}\hfill
\begin{minipage}[t]{0.355\linewidth}
\vspace{0pt}
\centering\footnotesize
\begin{minipage}[t][3\baselineskip][t]{\linewidth}
\caption{\textbf{Cross-view support and verification.} ( Submap $6\!\rightarrow\!60$; $16\!\rightarrow\!49$)}
\label{tab:ablation_connection_verification}
\end{minipage}\par
\fontsize{6}{7}\selectfont
\renewcommand{\arraystretch}{1.3}
\begin{tabular*}{\linewidth}{@{\extracolsep{\fill}}lp{0.34\linewidth}rrcl@{}}
\toprule
\shortstack{Submap\\pair} & \shortstack {Matching / \\validation} & \shortstack{Image\\pairs} & Inliers & \shortstack{Support\\A/B} & Test \\
\midrule
$6\!\rightarrow\!60$ & \raggedright Sampled & 16 & 58 & 1/1 & Fail \\
$6\!\rightarrow\!60$& \raggedright All frames & 256 & 1,843 & 10/14 & Pass \\
\midrule
$16\!\rightarrow\!49$ & \raggedright Sampled & 32 & 259 & 3/4 & Fail \\
$16\!\rightarrow\!49$ & \raggedright All frames / alternating groups & 256 & 3,990 & 6/14 & Fail \\
$16\!\rightarrow\!49$ & \raggedright All frames / balanced groups & 256 & 3,990 & 6/14 & Pass \\
\bottomrule
\end{tabular*}
\end{minipage}
\par\vspace{4pt}
\end{table}

\paragraph{Observation fusion and global refinement.}
Low reprojection error after local BA can coexist with global drift. PG reduces ATE while slightly increasing reprojection error, consistent with optimizing camera constraints before pixel-level refinement (Table~\ref{tab:ablation_staff_backend}, Fig.~\ref{fig:ablation_global_correction}). Fusion merges landmark tracks across revisits, increasing long-range support and changing the reprojection constraints. Subsequent BA reduces residuals and further improves trajectory accuracy.

\paragraph{Effect of depth scale calibration.}
Frame-specific calibration improves the precision--recall balance after registration, showing the value of accommodating scale variation within a prediction window (Table~\ref{tab:ablation_local_anchoring}). Whole-submap alignment retains higher recall. The advantage of per-frame calibration lies in the fidelity of the registered geometry and F-score.

\paragraph{Multi-view support and geometric verification.}
On the selected Residence2 pairs (small overlap long-range connections), all-frame matching increases support, while balanced groups allow verification to pass at unchanged thresholds (Table~\ref{tab:ablation_connection_verification}). Predicted depth expands 3D support, and pooling views improves correct recall.

\section{Conclusion}

In this paper, we study how local feed-forward geometry can support global consistency in long-sequence panoramic reconstruction. The local geometry of feed-forward model can provide 3D evidence for verifying distant feature matches and establishing camera constraints. The key insight is this mutual dependence: sparse mapping anchors local geometry, and local geometry strengthens the constraints used to correct the sparse map. Our seven-scene PanoRecon benchmark evaluates trajectory accuracy and reconstruction consistency under repetitive architecture and long-range revisits using reference poses and LiDAR geometry. Experiments across 360LOC, YUTO MMS, and PanoRecon show the improved mean ATE and high fidelity  dense reconstruction results.
\paragraph{Limitations and Future Work.}
Severe motion blur or prolonged low-texture segments can reduce the reliable feature observations needed for tracking and depth calibration. Our system also does not explicitly handle dynamic objects, such as moving vehicles and pedestrians, and does not yet achieve real-time reconstruction. Future work will focus on more robust correspondences and efficient feed-forward models for streaming reconstruction from $360^\circ$ video in dynamic scenes.

\bibliographystyle{plainnat}
\bibliography{main}

\appendix

\section{Additional Experiment Details}
\subsection{Datasets.}
\label{app:datasets}
\paragraph{360LOC.}
We use the four daytime mapping sequences of 360LOC~\citep{Huang_2024_CVPR}: Atrium, Concourse, Hall, and Piatrium. The released reference split contains 2,244 panoramas (491--632 per scene) at $6144\times3072$ resolution, with reference poses and LiDAR maps. Their indoor and outdoor settings, symmetric layouts, repeated structures, and moving pedestrians let us assess reconstruction consistency in visually ambiguous scenes.

For dense evaluation, we back-project the official per-ERP depth maps along spherical viewing rays and transform the resulting points into the world frame using the official camera poses. Valid points across mapping frames are merged by voxel averaging to form the reference point cloud.

\paragraph{YUTO MMS.}
We evaluate all four sequences (A--D) of YUTO MMS~\citep{Zhang_2024}, captured by a vehicle-mounted panoramic camera. The released sequences cover 20.1\,km and contain 24,360 panoramas. Sequence C alone spans 9.137\,km over 25 minutes with 10,778 images. The routes traverse campus and residential areas with large and small loops, frequent turns, and dynamic traffic. This scale and revisit structure let us evaluate accumulated drift and global consistency over extended trajectories, using reference poses obtained through offline GPS and LiDAR adjustment.

For dense evaluation, we construct the reference point cloud from the released LiDAR scans, ground-truth camera poses, and sensor calibration. We apply the calibrated LiDAR-to-camera transform, and accumulate the transformed scans in a common world frame.
\begin{figure}[!h]
    \centering
    \includegraphics[width=1\linewidth]{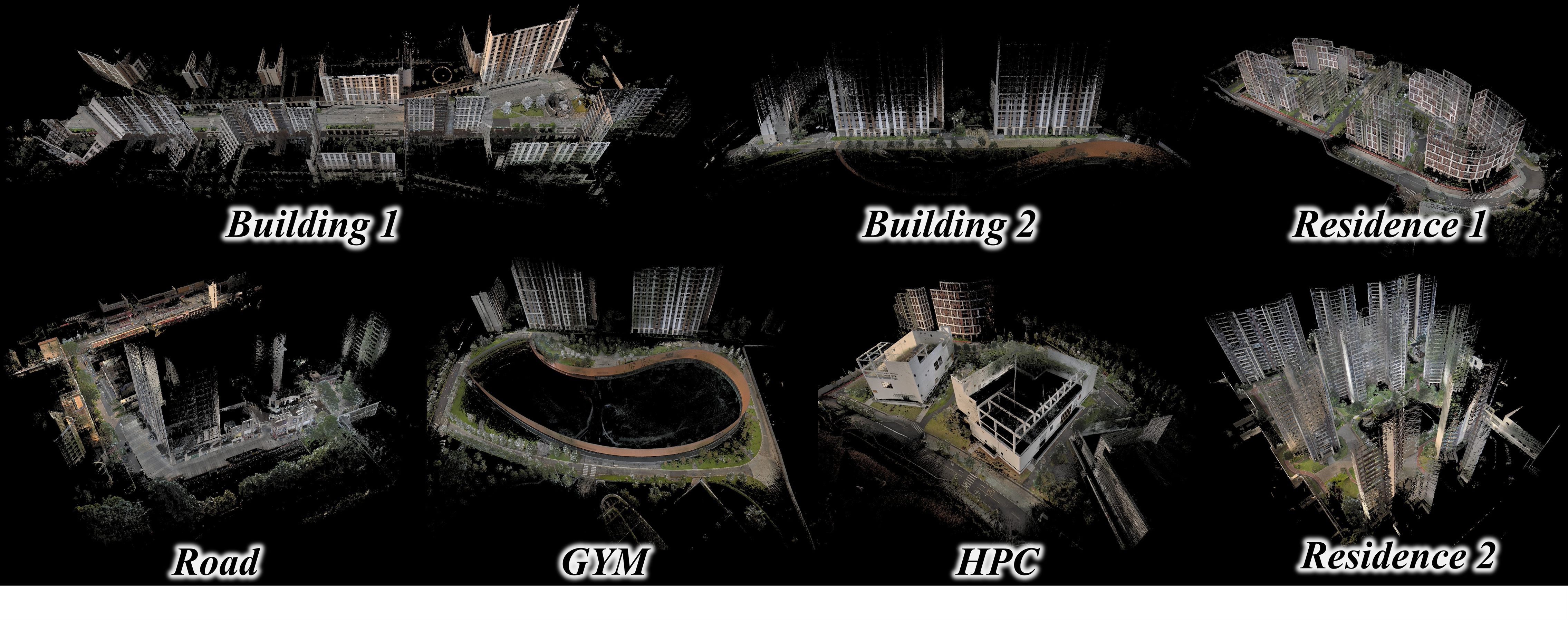}
    \caption{{Visualization of ours PanoRecon benchmark ground-truth LiDAR geometry.}}
    \label{fig:dataset_full.jpg}
\end{figure}
\paragraph{Our PanoRecon.}
\label{app:panorecon}
The PanoRecon benchmark evaluates global consistency under repeated architecture and limited overlap between distant revisits. We select seven scenes with low-texture surfaces and buildings that share similar geometry and appearance (Fig.~\ref{fig:dataset_full.jpg}). Each planned route contains a large loop, smaller loops, and repeated visits to the same buildings after long travel distances and time intervals. At revisits, the buildings are viewed from farther away or at oblique angles and occupy only a small region of the ERP image. The resulting limited overlap tests whether reconstruction methods can establish long-range connections and correct accumulated drift.

\textit{Data acquisition.} We independently acquire panoramic video and LiDAR data along the same planned route in each scene. The panoramic recordings use an Insta360 X4 at 5K resolution and 30\,fps. We mount the camera on a 3\,m extension pole to reduce the operator's presence in the ERP images. We export the recordings as ERP videos and sample 2K frames every 0.1\,s. The separate LiDAR survey uses RTK positioning and three onboard cameras for point-cloud colorization.

\textit{Data processing.} The two acquisition sessions are registered through reference images tied to the LiDAR coordinate system. We first post-process the LiDAR SLAM trajectory with RTK differential corrections and export the resulting point cloud, the three onboard cameras' poses, and their intrinsic and extrinsic calibration. We import these data into \textit{RealityScan} to render LiDAR reference images, whose camera poses remain fixed throughout registration.

Each ERP frame is split into perspective views with known intrinsics, a shared optical center, and fixed relative rotations. We import these views as a calibrated rig and globally match them to the fixed LiDAR reference images. Pose optimization updates only the panoramic rigs, with the reference cameras and rig calibration held fixed. We then convert the estimated perspective-view poses back to one ERP pose per frame using the rig extrinsics. This yields reference panoramic poses and LiDAR geometry in the same metric coordinate system.

\subsection{Baselines Settings.}
\label{app:baseline-settings}
All methods use 2K input images except PanoVGGT and Scal3R, which use their official default resizing settings.

\paragraph{Input sequences and evaluation frames.}
On 360LOC, SLAM methods process all mapping images and retain frame-wise poses for evaluation. On PanoRecon, SLAM methods use all images for localization and retain evaluation keyframes at 1\,s intervals. On YUTO MMS, SLAM methods process the full image sequences, and evaluation poses for both SLAM and feed-forward methods are sampled every three images. For these vehicle sequences with varying speeds, sampling follows frame indices. SfM methods use all 360LOC mapping images, sample PanoRecon inputs every 1\,s, and use every 3 images, following the corresponding evaluation sampling intervals.

\paragraph{Feed-forward inference.}
Feed-forward methods receive the full YUTO MMS sequences, except that Scal3R uses every third image on Sequences C and D, where full-sequence inputs cause GPU out-of-memory (OOM) errors. PanoVGGT encounters OOM on every evaluated scene with its official default resizing and our single-GPU setup.

\section{Additional Method Implementation Details}
\label{app:method-details}

\begin{figure}[!h]
    \centering
    \includegraphics[width=1\linewidth]{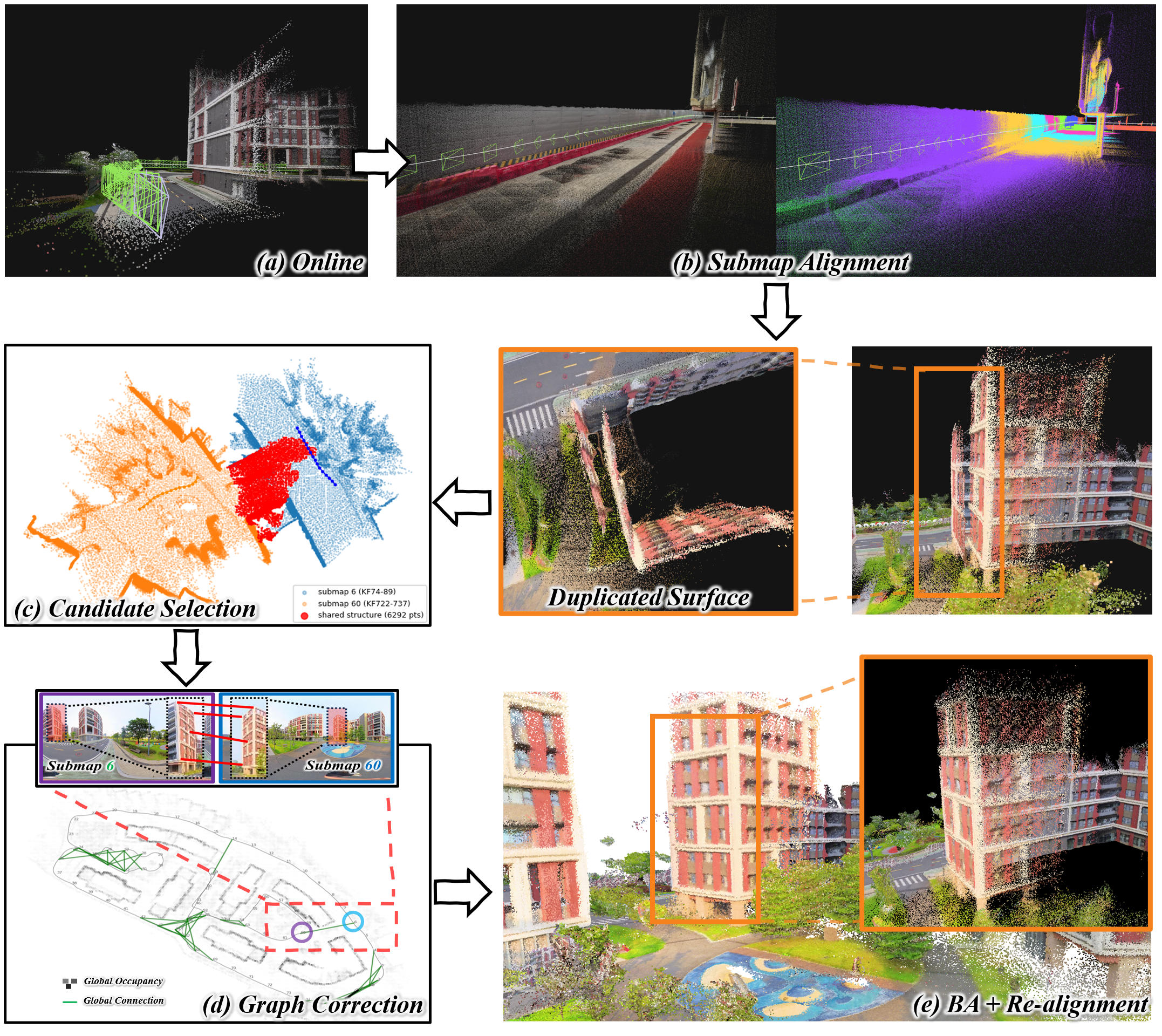}
    \caption{\textbf{From online submaps to globally consistent geometry.} (a) Online pose estimation and local reconstruction. (b) Submap alignment, with distinct colors indicating different submaps. (c) Candidate selection from geometric overlap. (d) Graph correction using verified long-range connections. (e) BA and realignment remove the duplicated building surfaces left by pose drift after online alignment.}
    \label{fig:flow}
\end{figure}

We provide implementation details for depth scale calibration, long-range connection verification, and global refinement, following the order and notation of Sec.~\ref{sec:method}. Figure~\ref{fig:flow} shows intermediate reconstructions and how global correction resolves duplicated building surfaces.
\subsection{Local Reconstruction Details}
\label{app:local-reconstruction}

\subsubsection{Sparse mapping and prediction windows}
The panoramic frontend tracks ERP images through projection matching and camera pose optimization (Fig.~\ref{fig:flow}(a)). New keyframes add observations and triangulated landmarks, which local BA refines before depth scale calibration.

We use the overlapping windows specified in Implementation. The first and last two inferred panoramas supply context for Pi3 and cached depths for connection matching. Scale fitting and point-cloud output use the interior frames.

We use the camera-to-world poses $T_i$ defined in Sec.~\ref{sec:method:calibration}, with the camera's $y$ axis pointing downward. For an ERP image of width $W$ and height $H$, pixel $\mathbf u=(u,v)$ has longitude $\theta=2\pi(u/W-1/2)$ and upward-positive latitude $\phi=\pi(1/2-v/H)$. Its unit bearing and the inverse projection are
\begin{equation}
\begin{aligned}
    \mathbf b(u,v)&=
       (\cos\phi\sin\theta,\;-\sin\phi,\;\cos\phi\cos\theta)^{\top},\\
    \pi_{\mathrm{ERP}}(\mathbf x)&=
       \begin{bmatrix}
       W\bigl(\tfrac12+\operatorname{atan2}(x,z)/(2\pi)\bigr)\\
       H\bigl(\tfrac12+\arcsin(y/\|\mathbf x\|_2)/\pi\bigr)
       \end{bmatrix},\quad \mathbf x=(x,y,z)^{\top}.
\end{aligned}
\label{eq:app:erp-model}
\end{equation}
The horizontal coordinate is periodic. Reconstruction distances use the sparse map's scale, with metric scale recovered through alignment for evaluation.

Each virtual perspective view has known intrinsics $K_v$, rotation $Q_v$ into the panoramic camera, and the same optical center as that camera. We take the optical-axis component $\hat z$ of each Pi3 local point and back-project it along the calibrated ray $K_v^{-1}\bar{\mathbf p}$. After scale calibration, its distance from the camera is $d=s_{wi}\hat z\|K_v^{-1}\bar{\mathbf p}\|_2$. We store the source keyframe, virtual view, pixel, and unscaled depth so that Eq.~(\ref{eq:method:dense-placement}) can update its world position after optimization.
\begin{figure}[h]
    \centering
    \includegraphics[width=1\linewidth]{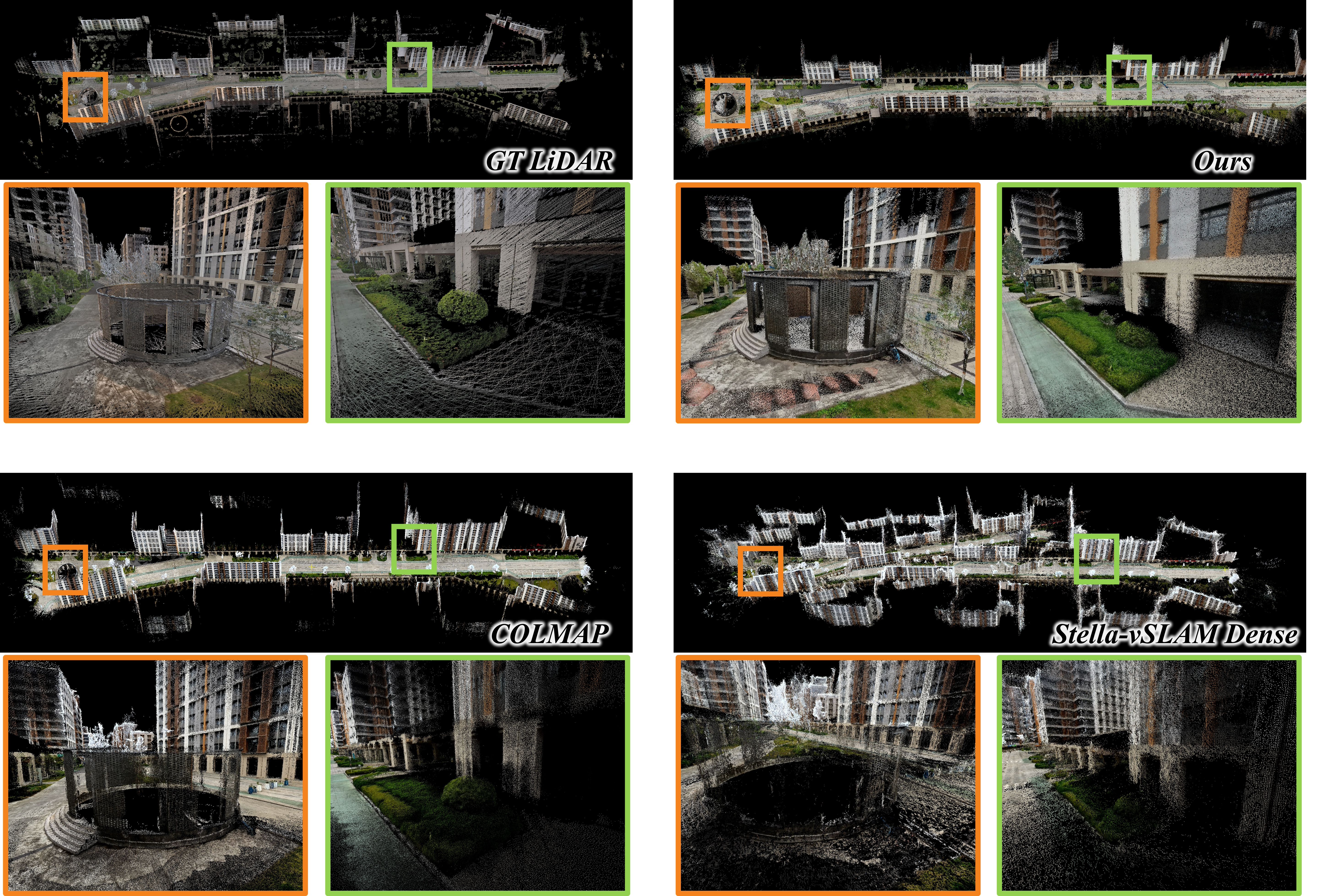}
    \caption{Point-cloud comparison of building facades and local structures in the PanoRecon benchmark's Building1 scene.}
    \label{fig:dense_building}
\end{figure}

\begin{figure}[t]
    \centering
    \includegraphics[width=1\linewidth]{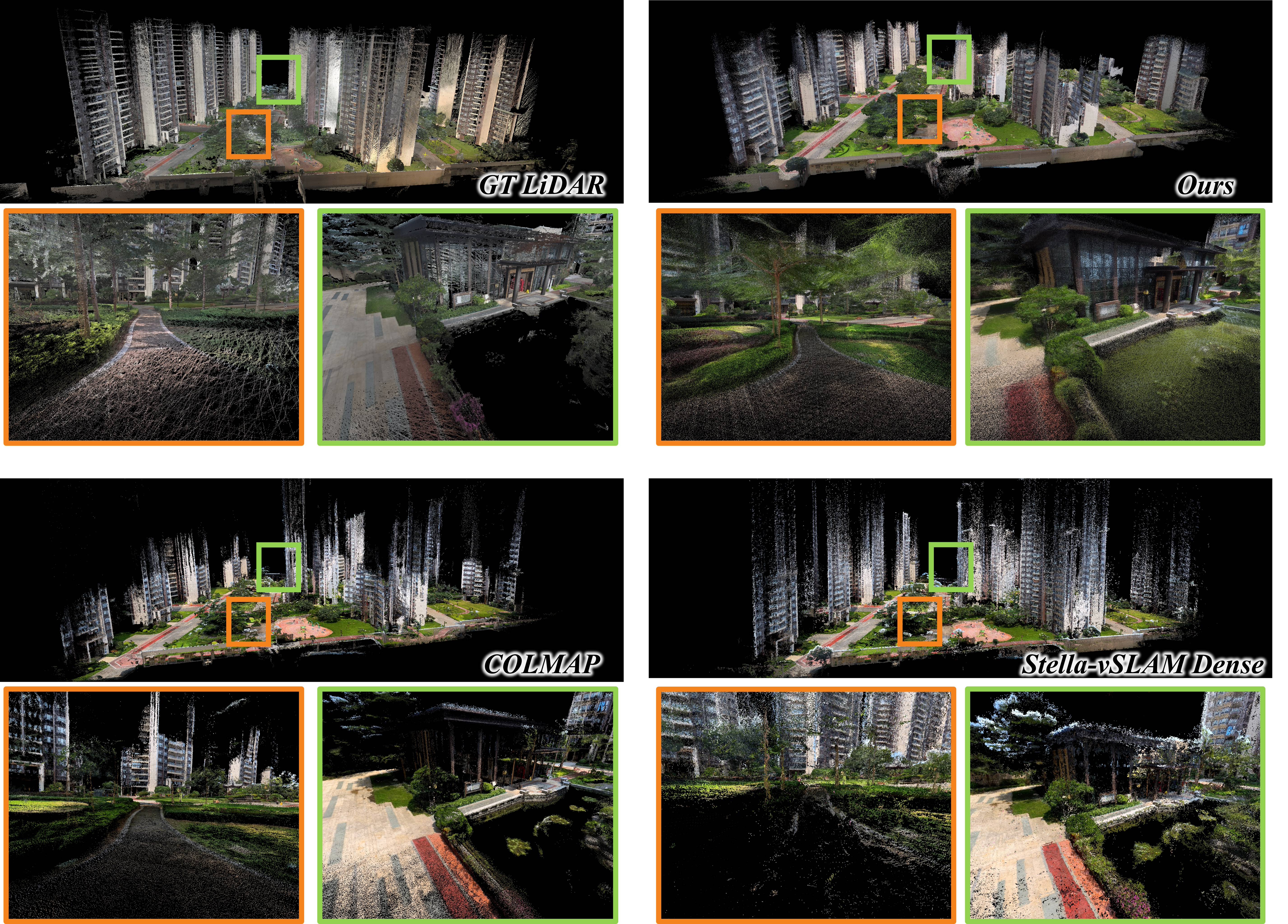}
    \caption{Point-cloud comparison in the Residence scene.}
    \label{fig:dense_residence}
\end{figure}
\subsubsection{Landmark observations and depth scale calibration}
\label{app:calibration}

For landmark $\mathbf L_j$ observed by keyframe $i$, the geometric depth in virtual view $v$ is
\begin{equation}
    z^{\mathrm{geom}}_{ijv}
       =\left[Q_v^{\top}R_i^{\top}(\mathbf L_j-\mathbf c_i)\right]_z,
\label{eq:app:anchor-depth}
\end{equation}
where $[\cdot]_z$ selects the optical-axis component. Comparing this depth with the prediction at the corresponding pixel gives the log ratio in Eq.~(\ref{eq:method:depth-calibration}). The fit estimates multiplicative depth scales, which are applied to the original predicted depths in Eq.~(\ref{eq:method:dense-placement}), producing the aligned submaps shown in Fig.~\ref{fig:flow}(b).

We filter landmark observations by positive depth, observation support, parallax, and reprojection error, and exclude masked pixels, view boundaries, and depth discontinuities. We repeat these checks against the updated sparse map before recalibration.

Before fitting, we filter log ratios by median absolute deviation. Frames with sufficient landmark support define $\mathcal A_w$, and windows with enough pooled observations are calibrated using Eq.~(\ref{eq:method:depth-calibration}). Temporal regularization links eligible frames that are adjacent in the sequence. The shared window scale initializes all frames, with $b_{wi}=0$, and frame corrections are fitted for eligible frames.

We solve Eq.~(\ref{eq:method:depth-calibration}) by iteratively reweighted least squares with $(\lambda_p,\lambda_s)=(32,64)$ and the Huber threshold specified in Implementation. The shared log scale $a_w$ starts from the median log ratio and is fitted without a bound. Frame corrections start at zero and are clipped after fitting to $\pm\log(1.05)$ relative to the shared scale.

Overlapping windows provide an additional scale adjustment from depths at common source keyframes, views, and pixels. Each shared keyframe contributes a median log-depth ratio, and their sample-count-weighted median gives the correction toward the preceding window. 
\begin{figure}[t]
    \centering
    \includegraphics[width=1\linewidth]{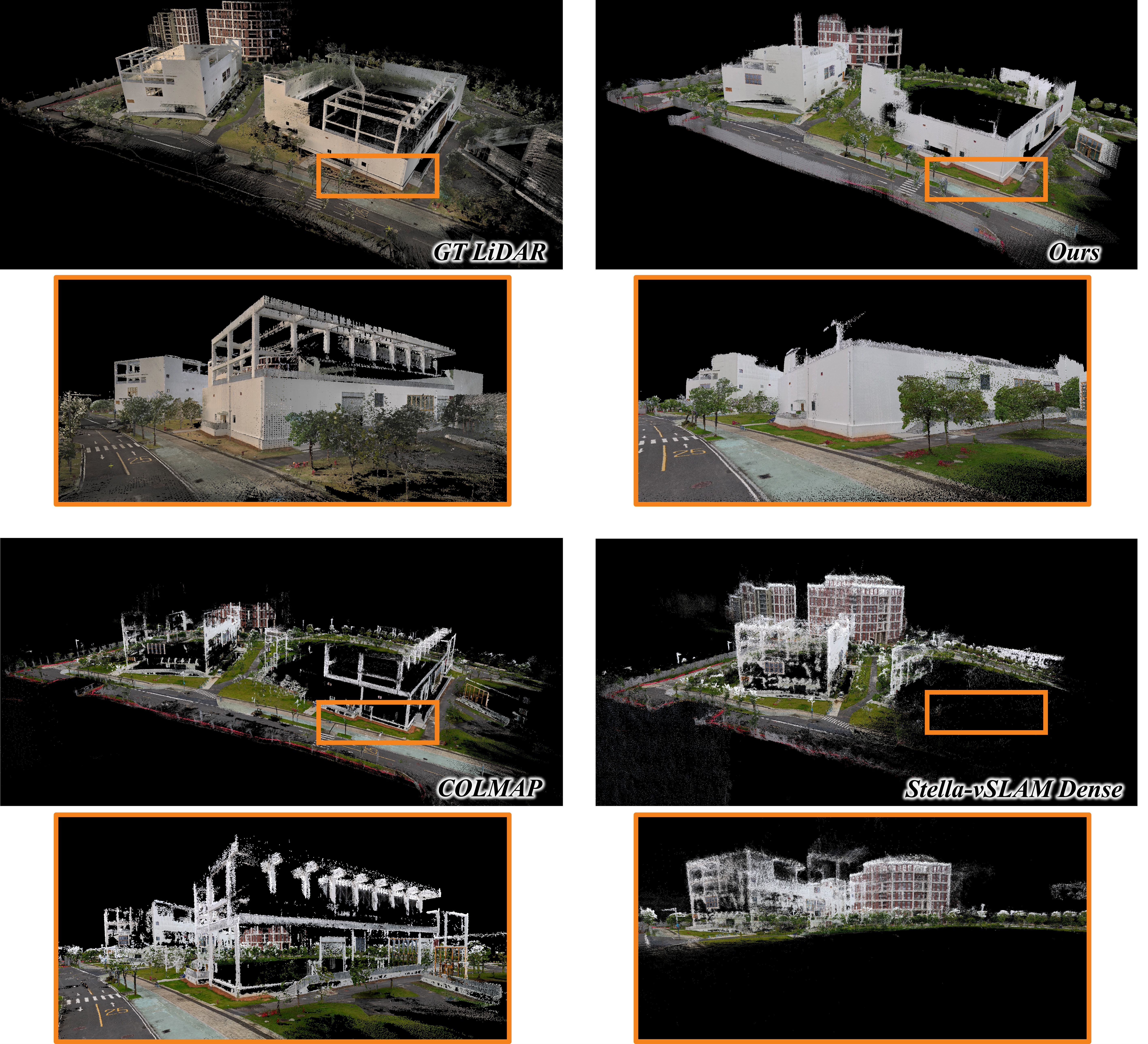}
    \caption{Point-cloud comparison in the HPC scene.}
    \label{fig:dense_hpc}
\end{figure}
\subsection{Global Connection Details}
\label{app:global-connections}

\subsubsection{Candidate submap selection}
\label{app:candidates}

We use calibrated submap points to restrict feature matching to spatial candidates (Fig.~\ref{fig:flow}(c)). For occupied voxel sets $V_A,V_B$ and one-voxel dilation $\mathcal D$, the overlap score is
\begin{equation}
    \eta_{AB}=\max\!\left\{
      \frac{|V_A\cap\mathcal D(V_B)|}{|V_A|},
      \frac{|V_B\cap\mathcal D(V_A)|}{|V_B|}
    \right\}.
\label{eq:app:overlap}
\end{equation}
Taking the larger fraction accommodates partial overlap between submaps of different sizes. We use voxel size $0.25 m$  in map units and require $\eta_{AB}\geq0.15$. A line-of-sight check tests for occlusion along rays from camera centers toward the overlapping region, requiring a mean visibility fraction of at least $0.6$.

To include nearby candidates displaced by drift, we expand robust axis-aligned submap bounds by 20\% on each side. Intersecting bounds supply up to three additional candidates. We also consider historical windows within two indices of selected candidates, adding at most 12 pairs per new window while retaining all original visibility candidates. Within this spatial neighborhood, candidate submaps separated by at least four window indices undergo the matching and verification below.

\subsubsection{Feature matching and geometric verification}
\label{app:correspondences}

We match all inferred keyframes with available poses and depths, including frames excluded from point-cloud output.  The median predicted camera distance in a local ERP neighborhood lifts each matched feature into 3D. Matches with valid depths on both sides form the correspondences in Sec.~\ref{sec:method:connections}.
\begin{figure}[t]
    \centering
    \includegraphics[width=1\linewidth]{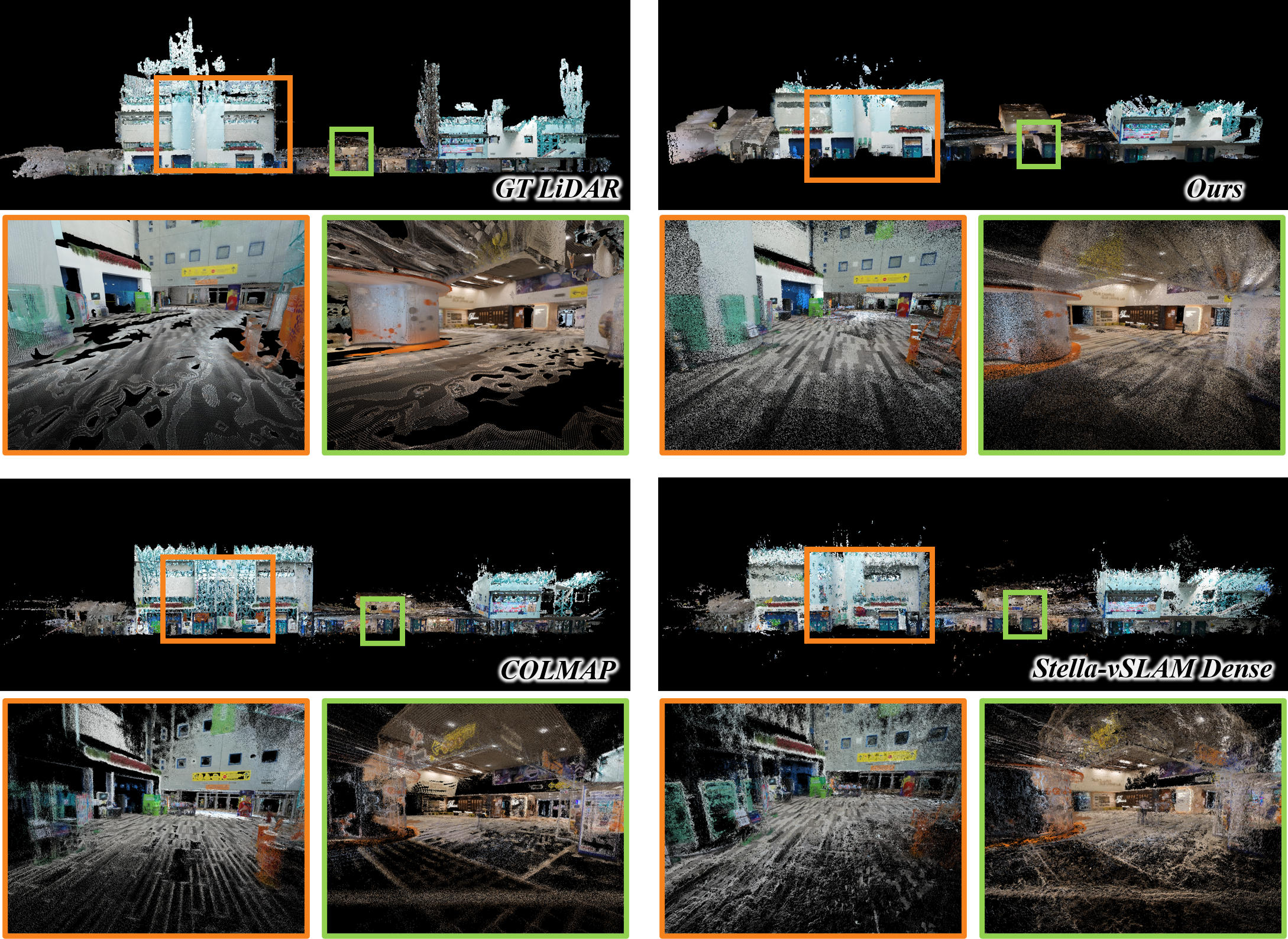}
    \caption{Point-cloud comparison in the 360LOC Concourse scene.}
    \label{fig:concourse}
\end{figure}

\begin{figure}[t]
    \centering
    \includegraphics[width=1\linewidth]{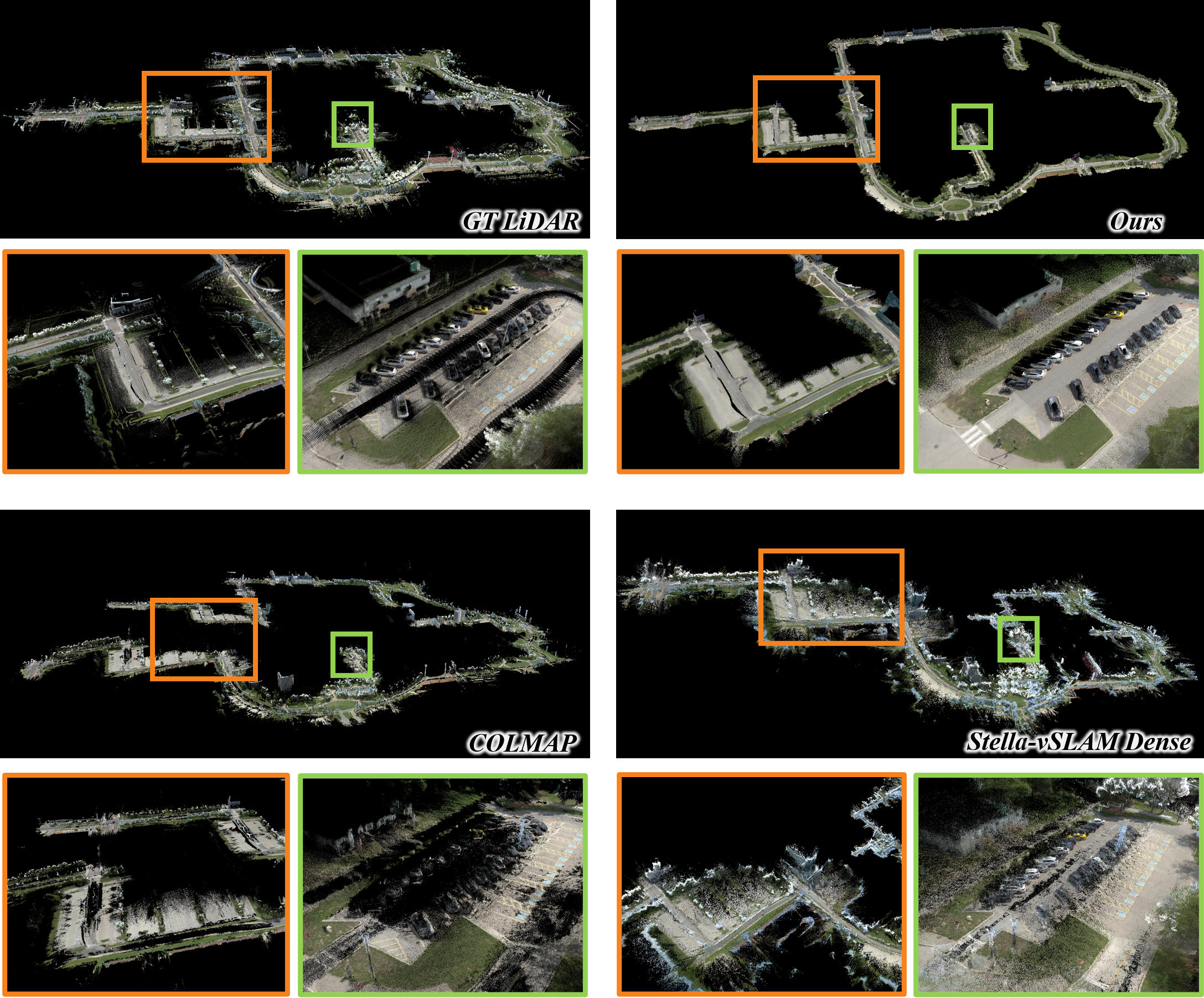}
    \caption{Point-cloud comparison in the YUTO MMS Sequence B scene.}
    \label{fig:sequenceb}
\end{figure}

\begin{figure}[t]
    \centering
    \includegraphics[width=1\linewidth]{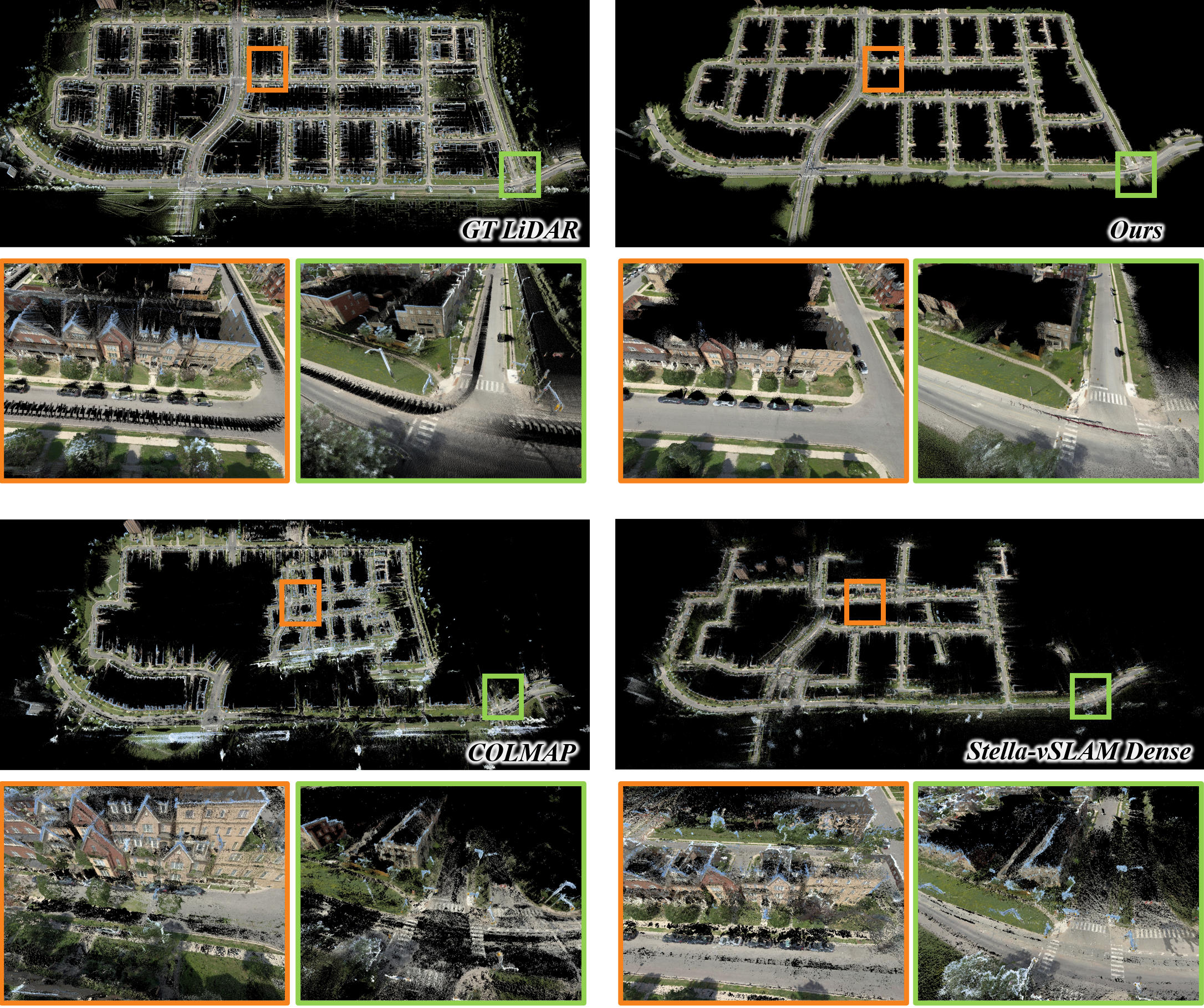}
    \caption{Point-cloud comparison in the YUTO MMS Sequence C scene.}
    \label{fig:sequencec}
\end{figure}
Each match retains its source keyframes and feature indices to distinguish repeated use of a feature from distinct feature support. We pool the 3D correspondences to estimate $S=S_{A\leftarrow B}$, with $S\mathbf x=sR\mathbf x+\mathbf t$, using RANSAC followed by least-squares refitting~\citep{umeyama1991transformation}.

For corresponding points $\mathbf P_a,\mathbf P_b$ and source camera centers $\mathbf c_a,\mathbf c_b$, geometric verification requires
\begin{equation}
\begin{aligned}
    \|S\mathbf P_b-\mathbf P_a\|_2&\leq\tau_d,\\
    \|S^{-1}\mathbf P_a-\mathbf P_b\|_2&\leq\tau_d,\\
    \angle(S\mathbf P_b-\mathbf c_a,\mathbf P_a-\mathbf c_a)
       &\leq\tau_\theta,\\
    \angle(S^{-1}\mathbf P_a-\mathbf c_b,\mathbf P_b-\mathbf c_b)
       &\leq\tau_\theta.
\end{aligned}
\label{eq:app:geometric-verification}
\end{equation}
Here $\angle(\mathbf x,\mathbf y)$ is the angle between nonzero vectors, and $\tau_d$ and $\tau_\theta$ are the distance and bearing thresholds specified in Implementation. The bidirectional distance test checks each submap at its own scale. We refit the similarity from the surviving correspondences and recompute inliers before the following checks.

\subsubsection{Validation across views and camera constraints}
\label{app:verification}

We require inliers from multiple views because repeated matches can refer to the same scene point. Table~\ref{tab:app:verification} specifies the required support, including distinct features and keyframes on each side. We reject nearly collinear inlier points on either side while allowing planar configurations.

Large corrections require additional validation using separate frame groups. Let $h$ be the median spacing between consecutive available keyframe centers, and let $\mathcal P_B$ contain the current submap's frames with predicted depths. A correction is large if its rotation exceeds $3^\circ$, its mean displacement of these camera centers exceeds $h$, or its scale lies outside $[0.95,1.05]$.

Before fitting, we divide the current frames into two similarly sized groups with balanced match counts. We estimate a transform from each group and verify it on the other group's matches. Each estimate $S_f=(s_f,R_f,\mathbf t_f)$ must also agree with the estimate from all matches, $S=(s,R,\mathbf t)$:
\begin{equation}
\begin{aligned}
    |\log(s_f/s)|&\leq0.05,\\
    \operatorname{angle}(R_fR^{\top})&\leq3^\circ,\\
    \max_{i\in\mathcal P_B}\|S_f\mathbf c_i-S\mathbf c_i\|_2
       &\leq\max(2h,2\tau_d).
\end{aligned}
\label{eq:app:fold-agreement}
\end{equation}
Here $\operatorname{angle}$ denotes the rotation angle. The groups use disjoint current frames and can share historical frames and scene content. The same transform-agreement criteria apply to verified historical submaps within two window indices that connect to the same current submap.

From each accepted submap pair, we select image pairs with at least ten inliers in descending count order. Their camera endpoints must be distinct on each side, and duplicate camera pairs are excluded across selected submap pairs. At most three camera constraints are formed using Eq.~(\ref{eq:method:camera-measurement}). For $n_{AB}$ inliers and $m_{AB}>0$ selected constraints, each constraint receives weight
\begin{equation}
    w_e=\frac{\operatorname{clip}(n_{AB}/3,10,100)}{m_{AB}}.
\label{eq:app:edge-weight}
\end{equation}
This normalization keeps each submap pair's total weight fixed across different numbers of selected camera edges.

\begin{table}[t]
\centering
\small
\caption{Connection-verification thresholds. Distances use map units, and $h$ is the median keyframe spacing.}
\label{tab:app:verification}
\begin{tabular}{p{0.51\linewidth}p{0.38\linewidth}}
\toprule
Quantity & Requirement \\
\midrule
Total inliers, small / large correction & $\geq60$ / $\geq100$ \\
Inlier fraction, small / large correction & $\geq0.35$ / $\geq0.50$ \\
Unique features per side & $\geq40$ \\
Supporting keyframes per side & $\geq3$, each with $\geq10$ inliers \\
Scale / rotation / mean camera displacement
    & $[2/3,1.5]$ / $\leq15^\circ$ / $\leq50h$ \\
Bidirectional distance / viewing angle & $\leq0.3$ / $\leq1^\circ$ \\
Estimation-group inliers & $\geq30$ \\
Validation-group inliers / fraction & $\geq30$ / $\geq0.40$ \\
Group and neighboring-transform agreement & Eq.~(\ref{eq:app:fold-agreement}) \\
\bottomrule
\end{tabular}
\end{table}

\subsubsection{Descriptor matching and cache reuse}
The CPU and CUDA implementations use the same mutual ORB matching rules and cache descriptors and feature matches. After global correction, we recompute 3D coordinates and geometric verification from the updated poses and depths.

\subsection{Global Refinement Details}
\label{app:global-refinement}

\subsubsection{Pose graph optimization and landmark fusion}
\label{app:refinement}

The pose graph combines spanning-tree edges, strong covisibility edges, and the new camera constraints from Sec.~\ref{sec:method:connections} to correct the trajectory (Fig.~\ref{fig:flow}(d)). Let $G_i\in\operatorname{Sim}(3)$ be a world-to-camera vertex, initialized from $T_i^{-1}$ with unit scale. For edge set $\mathcal E$, relative measurements $\widehat G_{b\leftarrow a}$, and weights $w_{ab}$, we fix the root vertex and minimize
\begin{equation}
    \min_{\{G_i\}}\sum_{(a,b)\in\mathcal E}
    w_{ab}\left\|
      \operatorname{Log}\!\left(\widehat G_{b\leftarrow a}G_aG_b^{-1}\right)
    \right\|_2^2.
    \label{eq:method:graph-objective}
\end{equation}
Here $\operatorname{Log}$ maps a similarity transform to its seven-dimensional tangent vector. Existing edges have unit weight, and new edges use Eq.~(\ref{eq:app:edge-weight}). After optimization, we check each new edge's relative error $H_{ab}=\widehat G_{b\leftarrow a}^{-1}G_bG_a^{-1}=(s_H,R_H,\mathbf t_H)$ against
\begin{equation}
    \operatorname{angle}(R_H)\leq5^\circ,
    \qquad \|\mathbf t_H/s_H\|_2\leq2,
    \qquad |\log s_H|\leq0.1.
\label{eq:app:graph-gate}
\end{equation}
The translation threshold uses map units. Pruning inconsistent new edges is followed by graph re-optimization and residual checks on the retained new constraints.

For each optimized world-to-camera similarity, dividing its translation by its scale gives the corresponding rigid transform. Inverting this transform yields the updated camera-to-world pose $T_i^+$. A landmark associated with reference keyframe $r(j)$ is updated by
\begin{equation}
    \mathbf L_j^+=(G_{r(j)}^+)^{-1}G_{r(j)}^0\mathbf L_j^0,
\label{eq:app:point-transport}
\end{equation}
where superscripts $0$ and $+$ denote the states before and after graph optimization.

We then apply the frontend's bidirectional projection matching between accepted camera endpoints and a limited set of their strongest covisible neighbors. Duplicate-landmark fusion and covisibility updates produce the shared observation set $\mathcal O$ for global BA.

\subsubsection{Panoramic bundle adjustment}
\label{app:gpu-ba}

We refine one rigid pose per panorama and the shared landmarks using the ERP reprojection objective in Eq.~(\ref{eq:method:erp-ba}). The root pose is fixed, while monocular scale remains unconstrained. The horizontal residual wraps to $[-W/2,W/2)$, and the Huber threshold is $\delta_b=\sqrt{5.99146}$.

\paragraph{Residuals for sparse automatic differentiation.}
We implement this objective in the BAE/PyPose framework~\citep{zhan2026bundle}. To express the robust loss as a sum of squared residuals, define $\mathbf e_{ij}=\sigma_{ij}^{-1}\mathbf r_{ij}$ and $q_{ij}=\|\mathbf e_{ij}\|^2$, and supply
\begin{equation}
    \mathbf f_{ij}=
    \begin{cases}
      \sqrt{\psi_{\delta_b}(q_{ij})/q_{ij}}\,\mathbf e_{ij},&q_{ij}>0,\\
      \mathbf 0,&q_{ij}=0.
    \end{cases}
    \qquad
    \sum_{(i,j)\in\mathcal O}\|\mathbf f_{ij}\|^2=E_{\mathrm{BA}}.
    \label{eq:app:gpu-robust-residual}
\end{equation}
The CPU and GPU backends optimize the same scalar objective. The GPU implementation differentiates the residual scaling, while g2o uses a robust Gauss--Newton approximation.

\paragraph{Matrix-free Schur optimization.}
Let $J_c$ and $J_p$ be the Jacobians of the stacked residual vector $\mathbf f$ with respect to free camera poses and landmarks. Diagonal damping gives camera and landmark blocks $U_\lambda$ and $V_\lambda$, with coupling $C=J_c^\top J_p$. For right-hand sides $\mathbf b_c=-J_c^\top\mathbf f$ and $\mathbf b_p=-J_p^\top\mathbf f$, eliminating landmark increments gives
\begin{equation}
\begin{aligned}
    S_\lambda\,\Delta\mathbf c
      &=\mathbf b_c-CV_\lambda^{-1}\mathbf b_p,
      &S_\lambda&=U_\lambda-CV_\lambda^{-1}C^\top,\\
    V_\lambda\,\Delta\mathbf p
      &=\mathbf b_p-C^\top\Delta\mathbf c.
\end{aligned}
\label{eq:app:gpu-schur}
\end{equation}
Here $\lambda$ denotes damping, and $\Delta\mathbf c$ and $\Delta\mathbf p$ stack camera pose and landmark increments. The reduced camera operator is evaluated through sparse products:
\begin{equation}
    S_\lambda\mathbf x=U_\lambda\mathbf x
      -J_c^\top\!\left[J_p\!\left(V_\lambda^{-1}
         \left[J_p^\top(J_c\mathbf x)\right]\right)\right].
    \label{eq:app:gpu-schur-product}
\end{equation}
This avoids storing $C$ or $S_\lambda$. The block-diagonal $V_\lambda$ has $3\times3$ blocks that can be inverted separately. GPU preconditioned conjugate gradients solve the camera system using a preconditioner derived from $U_\lambda$. The implementation also uses preconditioned conjugate gradients to recover landmark increments from the second equation in Eq.~(\ref{eq:app:gpu-schur}).

\paragraph{Solver settings and map updates.}
The GPU solver uses the precision and PCG settings in Implementation, with trust-region damping and at most 100 outer iterations. After at least two iterations, relative objective improvement below $10^{-3}$ stops refinement. Accepted steps have non-increasing objective values, and the final solution updates the camera poses and landmarks. The CPU backend uses analytic ERP Jacobians, g2o robust linearization, and a CSparse direct solver with OpenMP support. Initialization and local BA use the CPU backend, while the selected backend handles global BA.

\subsubsection{Dense reconstruction after global refinement}
\label{app:dense-update}

Updated poses and landmarks change both dense point positions and their depth references. For each affected window, we recheck its landmark observations, recompute Eq.~(\ref{eq:app:anchor-depth}), refit Eq.~(\ref{eq:method:depth-calibration}), and adjust overlapping predictions through their common samples. Equation~(\ref{eq:method:dense-placement}) then places the cached predictions using the updated poses and scales, yielding the realigned geometry in Fig.~\ref{fig:flow}(e).

At each submap boundary, accepted connections trigger graph correction, landmark fusion, and global BA before dense updates. Scheduled global BA also updates dense geometry. Frame input resumes once these operations finish.

\section{Additional Experiment Results}
\subsection{Runtime Analysis}
\label{app:runtime}
Table~\ref{tab:yuto-c-key-runtime} summarizes the instrumented reconstruction stages on YUTO MMS Sequence C. Our 257 Pi3 inference windows take 38.96 minutes, while 91 GPU BA solves take 0.87 minutes; connection matching and verification account for 56.60 minutes. In COLMAP, incremental reconstruction and BA take 618.52 minutes, and photometric and geometric MVS together take 609.85 minutes. We exclude separate image preparation, format conversion, and export. These historical runs use 10,778 frontend frames and 3,594 keyframes for our method, versus 3,593 sampled ERP images expanded into 43,116 perspective views for COLMAP. 
\begin{table}[!ht]
\centering\small
\setlength\tabcolsep{5pt}
\caption{\textbf{Key-stage runtime on YUTO MMS Sequence C.} Times are in minutes, excluding separate preprocessing and export.}
\label{tab:yuto-c-key-runtime}
\begin{tabular}{lr lr}
\toprule
\multicolumn{2}{c}{Ours} & \multicolumn{2}{c}{COLMAP (Pinhole)} \\
\cmidrule(lr){1-2}\cmidrule(lr){3-4}
Stage & Time & Stage & Time \\
\midrule
Frontend processing & 42.71 & Feature extraction & 7.21 \\
Pi3 inference & 38.96 & Matching and verification & 26.17 \\
Connection matching & 44.66 & Rig verification & 7.39 \\
Connection verification & 11.94 & Incremental reconstruction + BA & 618.52 \\
Pose-graph optimization & 10.08 & Photometric MVS & 227.33 \\
GPU BA solve & 0.87 & Geometric MVS & 382.52 \\
Post-BA realignment & 32.16 & Fusion & 261.72 \\
\midrule
Total & 181.38 & Total & 1530.86 \\
\bottomrule
\end{tabular}

\end{table}

\subsection{Applying Submap Alignment to PanoVGGT}
\label{app:panovggt-alignment}

PanoVGGT's joint inference exceeds GPU memory on the complete sequences in our evaluation, motivating a windowed reconstruction pipeline for long panoramic inputs. We therefore combine its panoramic predictions with the submap construction and retrieval workflow of VGGT-SLAM~\citep{guo2026panovggtfeedforward3dreconstruction,maggio2025vggtslamdensergbslam}. Since this integration is our own implementation, we present it separately in the supplement as a Residence2 case study, rather than as an official baseline. The experiment examines whether ours insight of using predicted geometry to align submaps and propagate global corrections can also benefit this reconstruction pipeline.

\paragraph{Re-implementation.}
We process Residence2 images at $518\times1036$ resolution in windows advancing by 16 images as submaps, with one image shared between consecutive windows. We then retrieval supplies at most one historical image for joint inference with each current window. PanoVGGT predicts camera poses and dense geometry directly from the ERP images. We express these predictions in the first camera's coordinate system and pass them to the VGGT-SLAM backend, retaining its sequential scale estimation and graph optimization. Valid points have finite coordinates and positive radial depth, since the panoramic checkpoint does not provide learned depth confidence. This produces the adapted-backend reference in Table~\ref{tab:panovggt-staff-alignment}.

\paragraph{Geometry-based submap alignment.}
We transfer our   submap alignment to this adapter using images shared across adjacent or retrieved windows. Corresponding pixels provide radial-depth ratios for robust relative scale estimation, while the predicted poses of the shared camera determine relative rotation and translation. These measurements form Sim(3) constraints for jointly optimizing submap poses and scales with a Huber loss, fixing the first submap. The optimized transformations update cached camera poses and dense points without rerunning PanoVGGT, propagating global corrections throughout the reconstruction.

\paragraph{Results.}
On Residence2, geometry-based submap alignment reduces ATE from 26.831 to 4.203\,m and raises F-score from 4.40\% to 18.67\%. Removing loop factors alone yields the lowest rotational RPE but substantially worse trajectory and reconstruction accuracy. 

\begin{table}[t]
\centering
\footnotesize
\setlength{\tabcolsep}{4pt}
\renewcommand{\arraystretch}{1.1}
\caption{\textbf{Adapt PanoVGGT with VGGT-SLAM on Residence2.}}
\label{tab:panovggt-staff-alignment}
\begin{tabular}{lrrrrrr}
\toprule
Configuration & ATE $\downarrow$ & RPE-t $\downarrow$ & RPE-r $\downarrow$ & Precision $\uparrow$ & Recall $\uparrow$ & F-score $\uparrow$ \\
 & (m) & (m) & ($^\circ$) & (\%) & (\%) & (\%) \\
\midrule
Default SL(4) & \multicolumn{6}{c}{Failed at submap 57/59} \\
Adapted SE(3) & 26.831 & 0.430 & 3.616 & 2.31 & 45.83 & 4.40 \\
Without loop factors SE(3) & 29.970 & 0.294 & \textbf{0.112} & 4.55 & 51.09 & 8.35 \\
Ours submap alignment Sim(3) & \textbf{4.203} & \textbf{0.107} & 0.149 & \textbf{10.60} & \textbf{78.14} & \textbf{18.67} \\
\bottomrule
\end{tabular}
\end{table}
\begin{figure}[!h]
    \centering
    \includegraphics[width=1\linewidth]{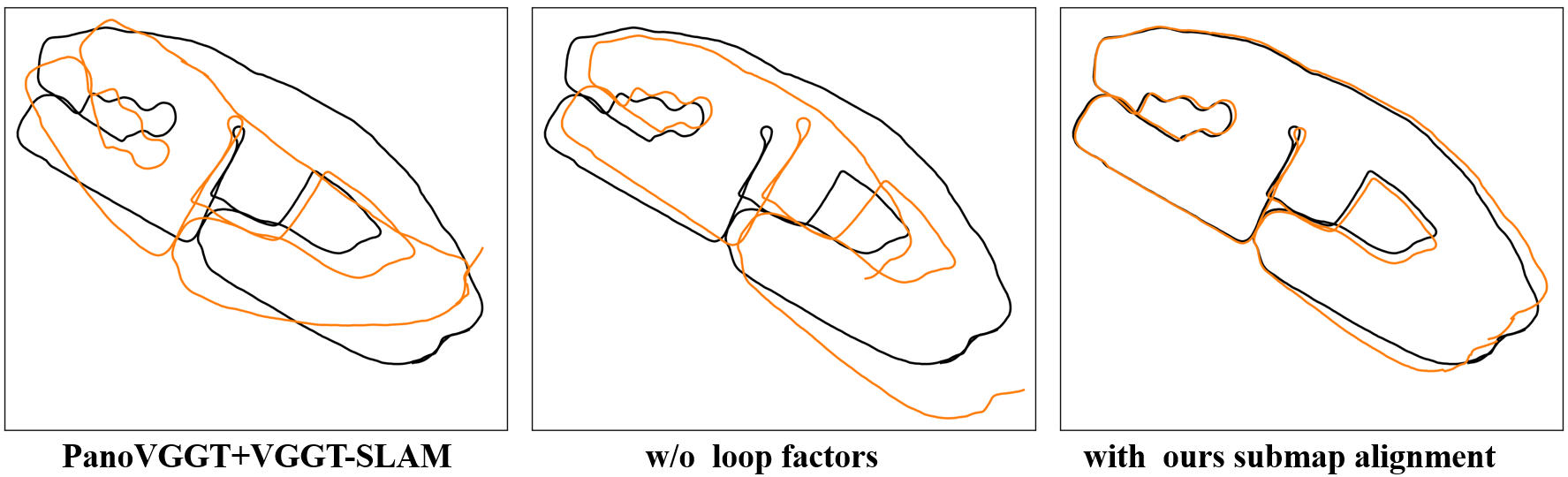}
    \caption{Applying Submap Alignment to PanoVGGT}
    \label{fig:panovggt}
\end{figure}

\begin{figure}[!h]
    \centering
    \includegraphics[width=1\linewidth]{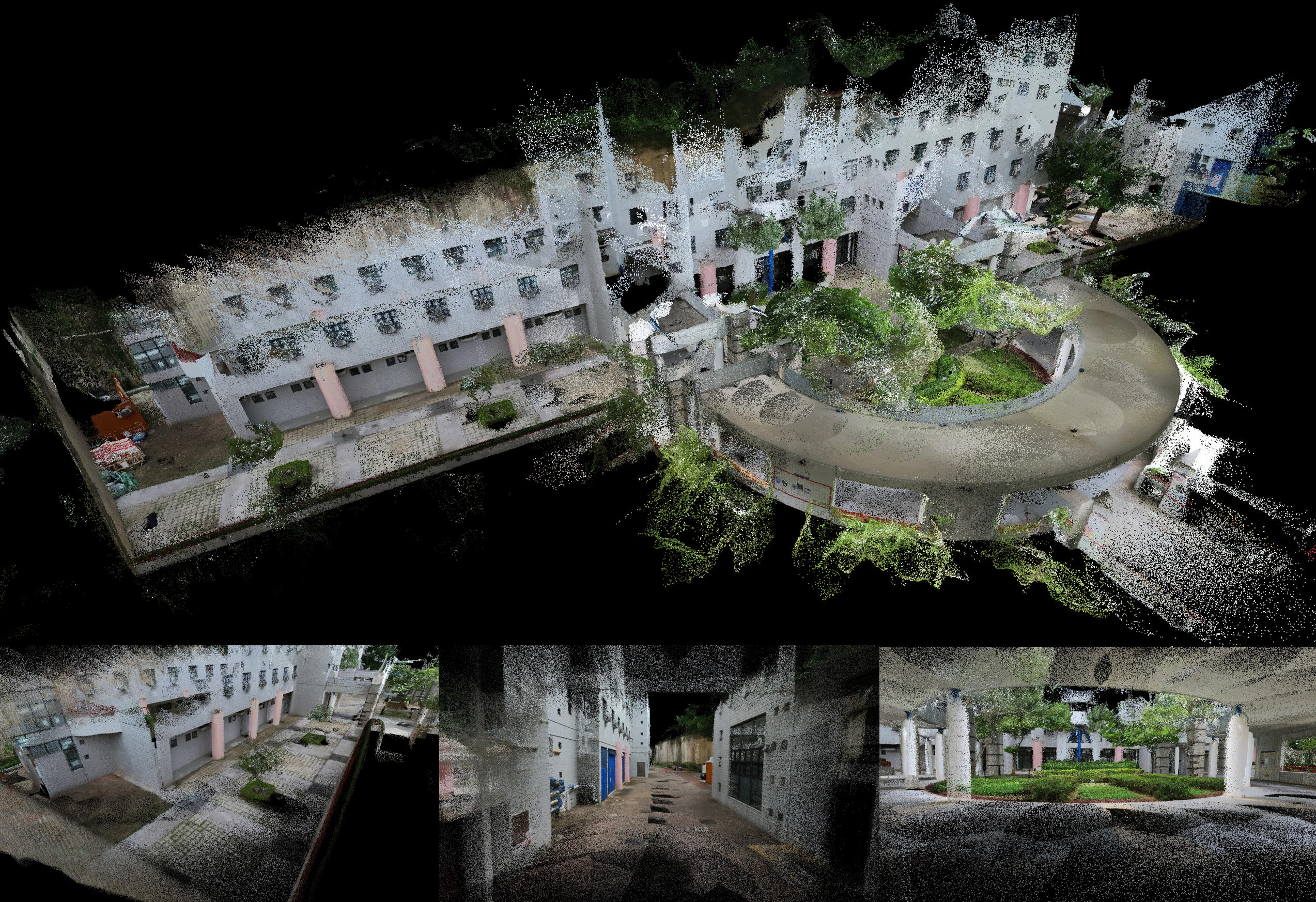}
    \caption{Reconstruction Result on 360loc Dataset Hall Scene.}
    \label{fig:hall}
\end{figure}

\begin{figure}[!h]
    \centering
    \includegraphics[width=1\linewidth]{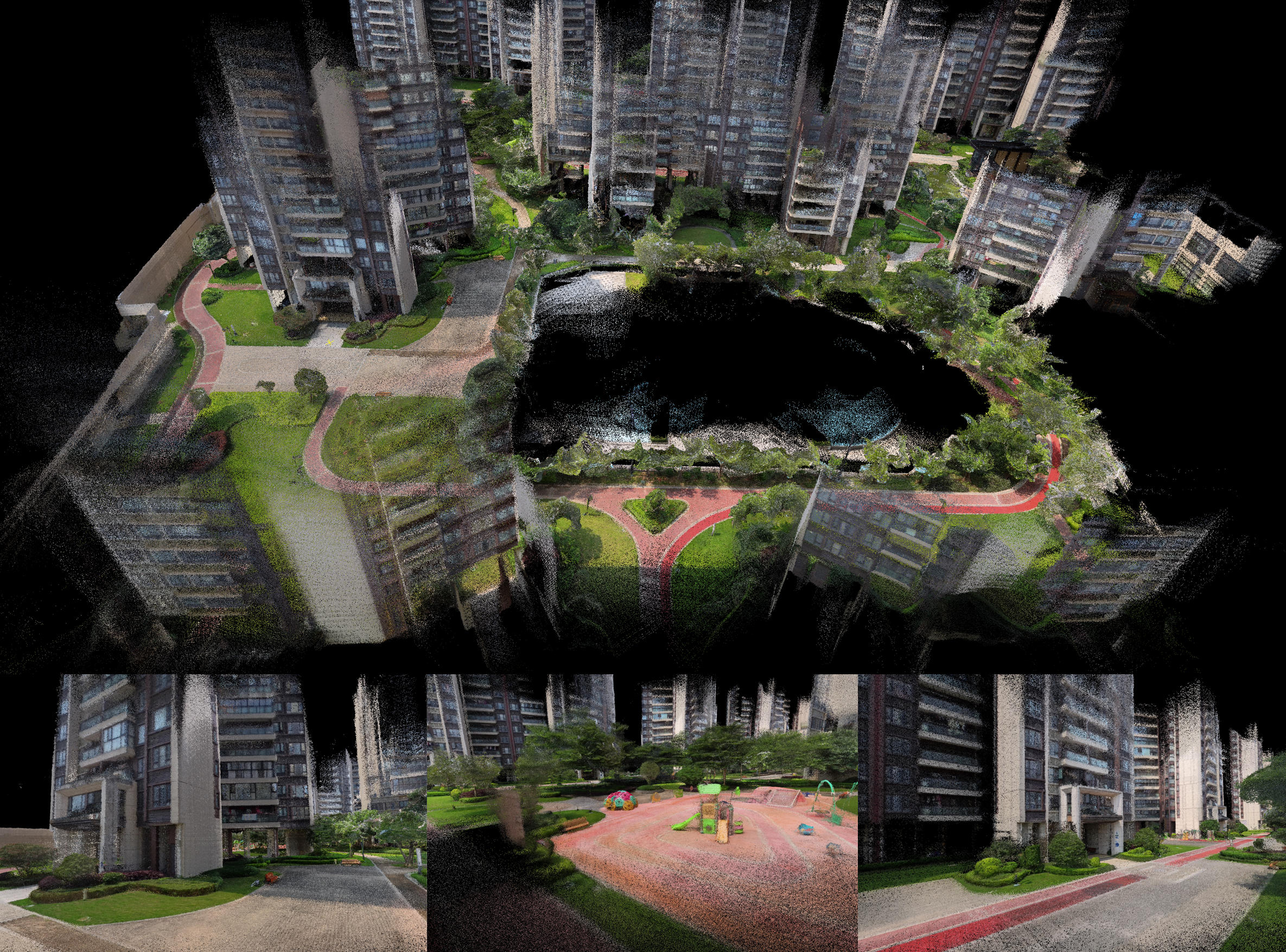}
    \caption{Reconstruction Result on Ours PanoRecon Benchmark Residence1 Scene.}
    \label{fig:res111}
\end{figure}

\begin{figure}[!h]
    \centering
    \includegraphics[width=1\linewidth]{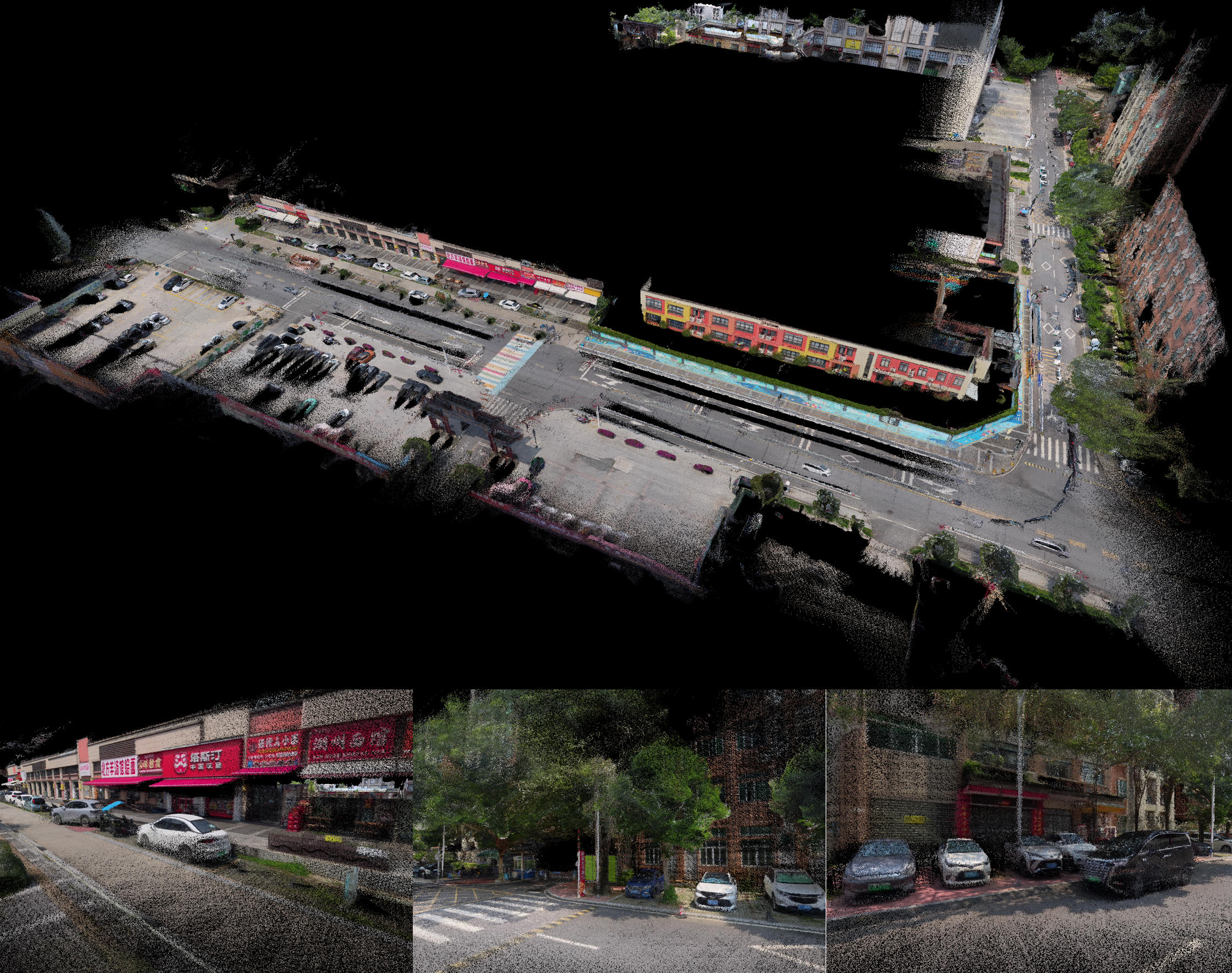}
    \caption{Reconstruction Result on Ours PanoRecon Benchmark Road Scene.}
    \label{fig:roadd}
\end{figure}

\begin{figure}[!h]
    \centering
    \includegraphics[width=1\linewidth]{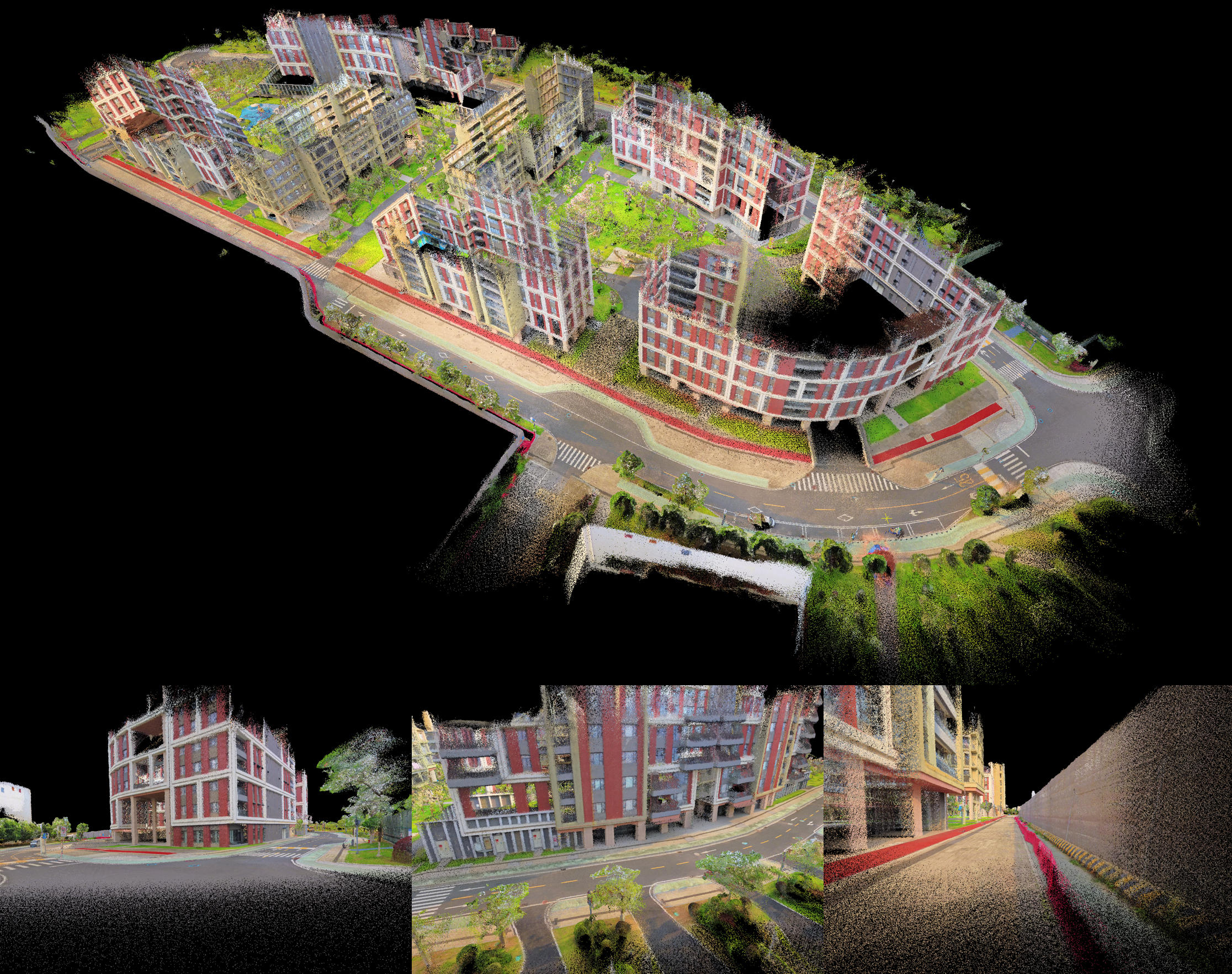}
    \caption{Reconstruction Result on Ours PanoRecon Benchmark Residence2 Scene.}
    \label{fig:staff}
\end{figure}

\begin{figure}[!h]
    \centering
    \includegraphics[width=1\linewidth]{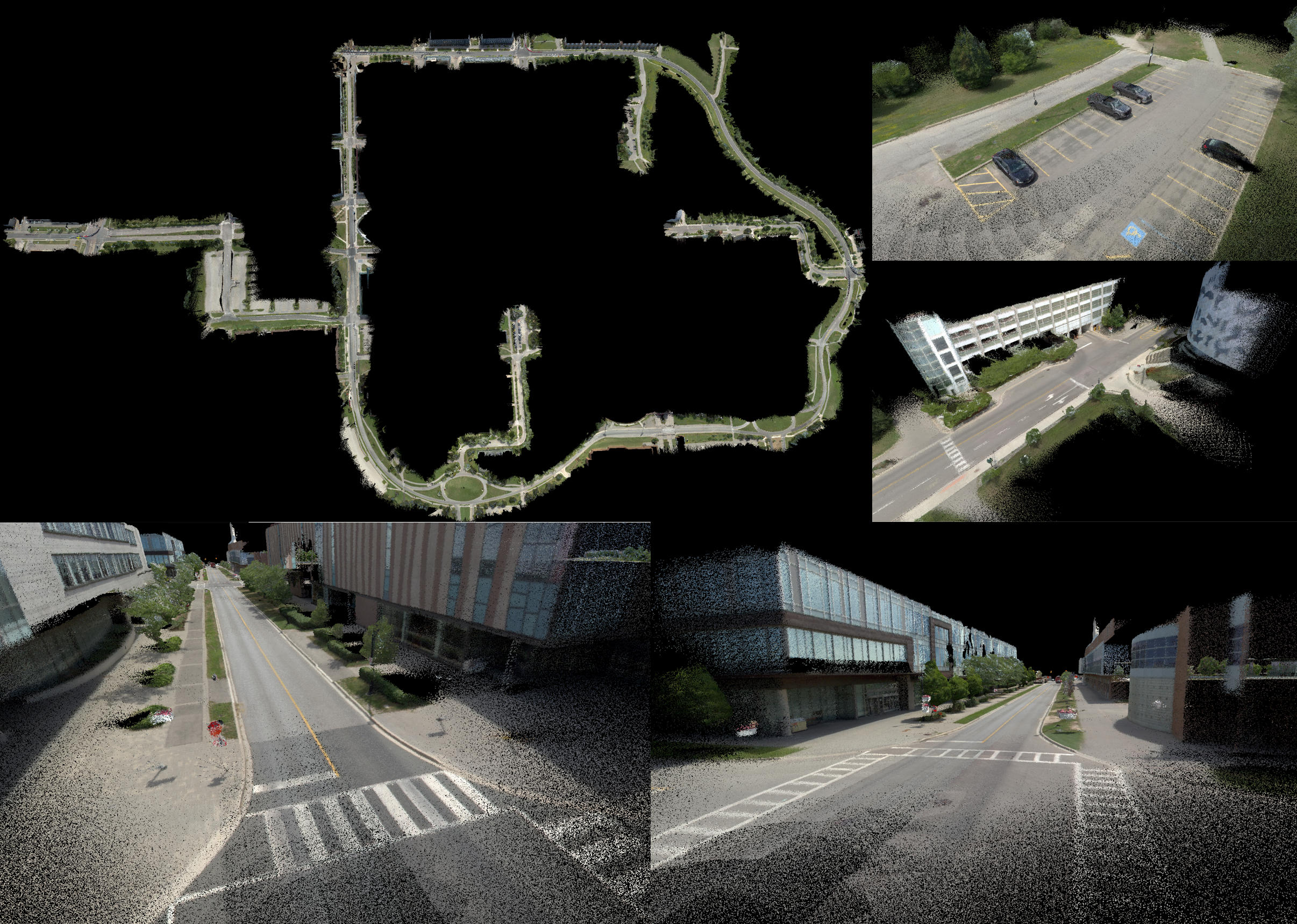}
    \caption{Reconstruction Result on YUTO MMS Benchmark Sequence B Scene.}
    \label{fig:seqbbb}
\end{figure}

\begin{figure}[!h]
    \centering
    \includegraphics[width=1\linewidth]{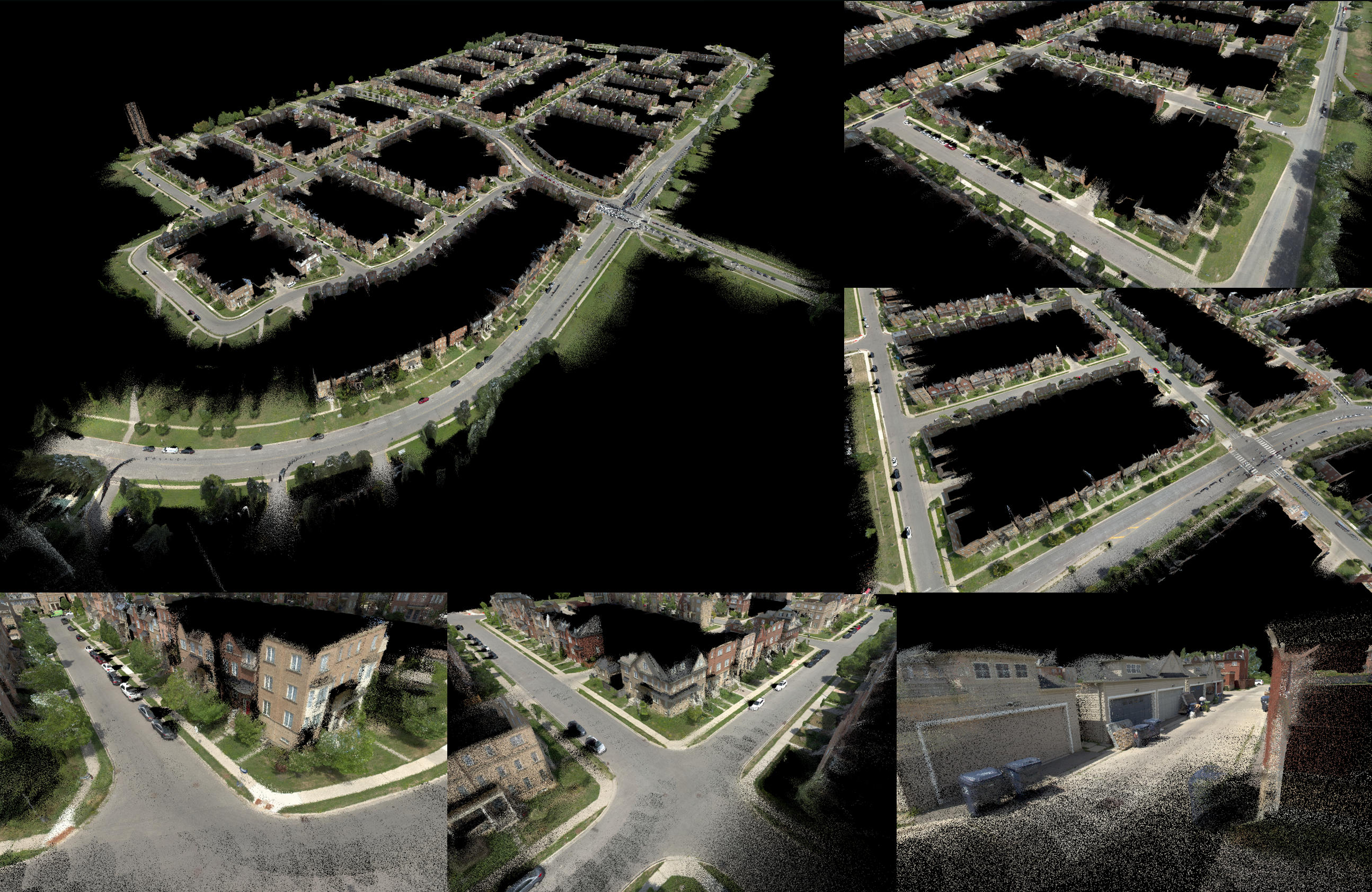}
    \caption{Reconstruction Result on YUTO MMS Benchmark Sequence C Scene.}
    \label{fig:seqccc}
\end{figure}

\section{Per-Scene Localization Results}
\label{app:localization-results}
Tables~\ref{tab:localization-360loc}, \ref{tab:localization-yuto}, and~\ref{tab:localization-panorecon} provide per-scene trajectory results for 360LOC, YUTO MMS, and PanoRecon, respectively.

\begin{table}[p]
\centering
\scriptsize
\setlength\tabcolsep{3pt}
\renewcommand\arraystretch{1.12}
\caption{\textbf{360LOC localization by scene.} ATE and RPE-t are in meters; RPE-r is in degrees. $n$ is the number of associated poses (scene panels) or available/expected scenes (Mean). All errors are RMSE; lower is better.}
\label{tab:localization-360loc}
\resizebox{\linewidth}{!}{
\begin{tabular}{l r r r r @{\hspace{12pt}} l r r r r}
\toprule
\multicolumn{5}{>{\columncolor{groupgray}}l}{\textbf{Atrium}} & \multicolumn{5}{>{\columncolor{groupgray}}l}{\textbf{Concourse}} \\
Method & $n$ & ATE$\downarrow$ & RPE-t$\downarrow$ & RPE-r$\downarrow$ & Method & $n$ & ATE$\downarrow$ & RPE-t$\downarrow$ & RPE-r$\downarrow$ \\
\midrule
Scal3R & 581 & 3.322 & 0.095 & 0.841 & Scal3R & 491 & 5.717 & 0.130 & 0.611 \\
360DVO & 581 & 0.654 & 0.044 & 0.034 & 360DVO & 491 & 0.779 & 0.011 & 0.044 \\
Open-vSLAM & 580 & 0.255 & 0.691 & 22.408 & Open-vSLAM & 471 & \first{0.081} & 0.670 & 20.111 \\
Stella-vSLAM & 285 & 1.209 & 0.083 & 0.059 & Stella-vSLAM & 262 & 0.172 & 0.037 & 0.298 \\
COLMAP (ERP) & 581 & \first{0.201} & \first{0.043} & \first{0.033} & COLMAP (ERP) & 491 & \second{0.131} & \second{0.007} & \second{0.038} \\
COLMAP (Pinhole) & 581 & \third{0.232} & \second{0.043} & \third{0.033} & COLMAP (Pinhole) & 491 & \third{0.159} & \first{0.007} & \first{0.037} \\
Spherical-SfM (ERP) & 581 & 0.354 & \third{0.043} & \second{0.033} & Spherical-SfM (ERP) & 491 & 0.211 & \third{0.010} & 0.042 \\
\textbf{Ours} & 581 & \second{0.208} & 0.062 & 0.040 & \textbf{Ours} & 491 & 0.169 & 0.016 & \third{0.040} \\
\midrule
\multicolumn{5}{>{\columncolor{groupgray}}l}{\textbf{Hall}} & \multicolumn{5}{>{\columncolor{groupgray}}l}{\textbf{Piatrium}} \\
Method & $n$ & ATE$\downarrow$ & RPE-t$\downarrow$ & RPE-r$\downarrow$ & Method & $n$ & ATE$\downarrow$ & RPE-t$\downarrow$ & RPE-r$\downarrow$ \\
\midrule
Scal3R & 540 & 11.008 & 0.172 & 3.758 & Scal3R & 632 & 1.417 & 0.099 & 0.205 \\
360DVO & 540 & 0.921 & 0.028 & \first{0.039} & 360DVO & 632 & 1.460 & 0.021 & 0.038 \\
Open-vSLAM & 540 & \second{0.714} & 0.721 & 23.225 & Open-vSLAM & 612 & 0.713 & 0.762 & 22.993 \\
Stella-vSLAM & 363 & \third{0.793} & 0.035 & 0.064 & Stella-vSLAM & 350 & 0.652 & 0.033 & 0.167 \\
COLMAP (ERP) & 540 & 0.919 & \second{0.018} & \third{0.041} & COLMAP (ERP) & 632 & \first{0.170} & \first{0.013} & \first{0.035} \\
COLMAP (Pinhole) & 540 & 1.027 & \third{0.021} & \second{0.041} & COLMAP (Pinhole) & 632 & \second{0.177} & \second{0.013} & \second{0.035} \\
Spherical-SfM (ERP) & -- & 1.093 & 0.027 & 0.042 & Spherical-SfM (ERP) & 540 & \third{0.308} & \third{0.015} & \third{0.036} \\
\textbf{Ours} & 540 & \first{0.218} & \first{0.015} & 0.054 & \textbf{Ours} & 632 & 0.484 & 0.027 & 0.041 \\
\midrule
\multicolumn{5}{>{\columncolor{groupgray}}l}{\textbf{Mean}} & \multicolumn{5}{>{\columncolor{groupgray}}l}{\textbf{}} \\
Method & $n$ & ATE$\downarrow$ & RPE-t$\downarrow$ & RPE-r$\downarrow$ &  & & & &  \\
\midrule
Scal3R & 4/4 & 5.366 & 0.124 & 1.354 &  &  &  &  &  \\
360DVO & 4/4 & 0.954 & 0.026 & 0.039 &  &  &  &  &  \\
Open-vSLAM & 4/4 & 0.441 & 0.711 & 22.184 &  &  &  &  &  \\
Stella-vSLAM & 4/4 & 0.706 & 0.047 & 0.147 &  &  &  &  &  \\
COLMAP (ERP) & 4/4 & \second{0.355} & \first{0.020} & \second{0.037} &  &  &  &  &  \\
COLMAP (Pinhole) & 4/4 & \third{0.399} & \second{0.021} & \first{0.037} &  &  &  &  &  \\
Spherical-SfM (ERP) & 4/4 & 0.491 & \third{0.024} & \third{0.038} &  &  &  &  &  \\
\textbf{Ours} & 4/4 & \first{0.270} & 0.030 & 0.044 &  &  &  &  &  \\
\bottomrule
\end{tabular}}
\end{table}

\begin{table}[p]
\centering
\scriptsize
\setlength\tabcolsep{3pt}
\renewcommand\arraystretch{1.12}
\caption{\textbf{YUTO MMS localization by scene.} ATE and RPE-t are in meters; RPE-r is in degrees. $n$ is the number of KF poses or available/expected scenes (Mean).}
\label{tab:localization-yuto}
\resizebox{\linewidth}{!}{
\begin{tabular}{l r r r r @{\hspace{12pt}} l r r r r}
\toprule
\multicolumn{5}{>{\columncolor{groupgray}}l}{\textbf{Sequence A}} & \multicolumn{5}{>{\columncolor{groupgray}}l}{\textbf{Sequence B}} \\
Method & $n$ & ATE$\downarrow$ & RPE-t$\downarrow$ & RPE-r$\downarrow$ & Method & $n$ & ATE$\downarrow$ & RPE-t$\downarrow$ & RPE-r$\downarrow$ \\
\midrule
Scal3R & 234 & 24.591 & 0.301 & 0.409 & Scal3R & 2794 & 307.710 & \third{0.948} & 0.381 \\
360DVO & 234 & \third{0.244} & \first{0.006} & \first{0.010} & 360DVO & 2794 & 135.769 & \second{0.277} & \first{0.013} \\
Open-vSLAM & 235 & 35.215 & 15.562 & 56.725 & Open-vSLAM & 2794 & 479.963 & 37.245 & 1.025 \\
Stella-vSLAM & 235 & 0.370 & 0.026 & 0.060 & Stella-vSLAM & 2794 & 252.571 & 1.074 & 0.223 \\
COLMAP (ERP) & 234 & \second{0.161} & \second{0.008} & \second{0.014} & COLMAP (ERP) & 2794 & 412.900 & 2.060 & 0.197 \\
COLMAP (Pinhole) & 234 & 0.474 & 0.017 & \third{0.017} & COLMAP (Pinhole) & 2794 & \third{75.064} & 0.954 & \third{0.036} \\
Spherical-SfM (ERP) & 234 & 0.409 & 2.036 & 4.139 & Spherical-SfM (ERP) & 2794 & \second{26.071} & 4.008 & 3.960 \\
\textbf{Ours} & 235 & \first{0.082} & \third{0.009} & 0.018 & \textbf{Ours} & 2795 & \first{3.367} & \first{0.041} & \second{0.024} \\
\midrule
\multicolumn{5}{>{\columncolor{groupgray}}l}{\textbf{Sequence C}} & \multicolumn{5}{>{\columncolor{groupgray}}l}{\textbf{Sequence D}} \\
Method & $n$ & ATE$\downarrow$ & RPE-t$\downarrow$ & RPE-r$\downarrow$ & Method & $n$ & ATE$\downarrow$ & RPE-t$\downarrow$ & RPE-r$\downarrow$ \\
\midrule
Scal3R & 3593 & 236.298 & 9.695 & 3.007 & Scal3R & 1500 & 181.028 & 0.700 & 0.302 \\
360DVO & 3593 & 115.432 & \second{0.366} & \first{0.011} & 360DVO & 1500 & 21.804 & \first{0.083} & \first{0.011} \\
Open-vSLAM & 3593 & 280.763 & 12.037 & 6.752 & Open-vSLAM & 1500 & 331.233 & 101.917 & 85.795 \\
Stella-vSLAM & 3593 & \second{12.848} & \third{0.714} & 0.166 & Stella-vSLAM & 1500 & 62.904 & 0.812 & 0.076 \\
COLMAP (ERP) & 3593 & 73.706 & 1.017 & \third{0.028} & COLMAP (ERP) & 1500 & \third{11.193} & \third{0.170} & \third{0.024} \\
COLMAP (Pinhole) & 3593 & 124.179 & 1.097 & 0.039 & COLMAP (Pinhole) & 1500 & 253.552 & 101.730 & 4.090 \\
Spherical-SfM (ERP) & 3593 & \third{54.137} & 3.858 & 4.076 & Spherical-SfM (ERP) & 1500 & \first{4.128} & 3.346 & 3.781 \\
\textbf{Ours} & 3594 & \first{2.474} & \first{0.068} & \second{0.022} & \textbf{Ours} & 1501 & \second{8.184} & \second{0.090} & \second{0.022} \\
\midrule
\multicolumn{5}{>{\columncolor{groupgray}}l}{\textbf{Mean}} & \multicolumn{5}{>{\columncolor{groupgray}}l}{\textbf{}} \\
Method & $n$ & ATE$\downarrow$ & RPE-t$\downarrow$ & RPE-r$\downarrow$ &  & & & &  \\
\midrule
Scal3R & 4/4 & 187.407 & 2.911 & 1.025 &  &  &  &  &  \\
360DVO & 4/4 & \third{68.313} & \second{0.183} & \first{0.011} &  &  &  &  &  \\
Open-vSLAM & 4/4 & 281.793 & 41.690 & 37.574 &  &  &  &  &  \\
Stella-vSLAM & 4/4 & 82.173 & \third{0.656} & 0.131 &  &  &  &  &  \\
COLMAP (ERP) & 4/4 & 124.490 & 0.814 & \third{0.066} &  &  &  &  &  \\
COLMAP (Pinhole) & 4/4 & 113.317 & 25.949 & 1.046 &  &  &  &  &  \\
Spherical-SfM (ERP) & 4/4 & \second{21.186} & 3.312 & 3.989 &  &  &  &  &  \\
\textbf{Ours} & 4/4 & \first{3.527} & \first{0.052} & \second{0.022} &  &  &  &  &  \\
\bottomrule
\end{tabular}}
\end{table}

\begin{table}[p]
\centering
\scriptsize
\setlength\tabcolsep{3pt}
\renewcommand\arraystretch{1.12}
\caption{\textbf{PanoRecon localization by scene.} ATE and RPE-t are in meters; RPE-r is in degrees. $n$ is the number of associated poses (scene panels) or available/expected scenes (Mean). All errors are RMSE; lower is better.}
\label{tab:localization-panorecon}
\resizebox{\linewidth}{!}{
\begin{tabular}{l r r r r @{\hspace{12pt}} l r r r r}
\toprule
\multicolumn{5}{>{\columncolor{groupgray}}l}{\textbf{Building 1}} & \multicolumn{5}{>{\columncolor{groupgray}}l}{\textbf{Building 2}} \\
Method & $n$ & ATE$\downarrow$ & RPE-t$\downarrow$ & RPE-r$\downarrow$ & Method & $n$ & ATE$\downarrow$ & RPE-t$\downarrow$ & RPE-r$\downarrow$ \\
\midrule
Scal3R & 765 & 14.643 & 0.281 & 0.182 & Scal3R & 419 & \first{3.537} & 0.223 & 0.128 \\
360DVO & 765 & 2.726 & \second{0.240} & \first{0.023} & 360DVO & 419 & \second{4.548} & \first{0.175} & \second{0.047} \\
Open-vSLAM & 765 & 3.387 & 2.836 & 34.833 & Open-vSLAM & 418 & 5.764 & 3.636 & 0.135 \\
Stella-vSLAM & 765 & \third{2.506} & 0.411 & 0.646 & Stella-vSLAM &  418 & \third{5.018} & 1.024 & 0.121 \\
COLMAP (ERP) & 765 & 10.831 & 15.002 & 8.499 & COLMAP (ERP) & 419 & 5.369 & \third{0.202} & \third{0.048} \\
COLMAP (Pinhole) & 765 & \first{1.618} & \third{0.240} & \third{0.026} & COLMAP (Pinhole) & 419 & 5.431 & 0.205 & \first{0.045} \\
Spherical-SfM (ERP) &765 & 11.907 & 15.817 & 8.852 & Spherical-SfM (ERP) & 419 & 30.653 & 25.932 & 0.059 \\
\textbf{Ours} & 764 & \second{1.620} & \first{0.240} & \second{0.024} & \textbf{Ours} & 418 & 5.302 & \second{0.201} & 0.052 \\
\midrule
\multicolumn{5}{>{\columncolor{groupgray}}l}{\textbf{Gym}} & \multicolumn{5}{>{\columncolor{groupgray}}l}{\textbf{HPC}} \\
Method & $n$ & ATE$\downarrow$ & RPE-t$\downarrow$ & RPE-r$\downarrow$ & Method & $n$ & ATE$\downarrow$ & RPE-t$\downarrow$ & RPE-r$\downarrow$ \\
\midrule
Scal3R & 480 & 8.943 & 0.533 & 0.501 & Scal3R & 507 & 1.381 & 0.239 & 0.533 \\
360DVO & 480 & 1.876 & \third{0.037} & \third{0.023} & 360DVO & 507 & 1.288 & 0.123 & 0.080 \\
Open-vSLAM & 480 & 17.622 & 13.582 & 56.856 & Open-vSLAM & 507 & \first{0.673} & 10.790 & 0.224 \\
Stella-vSLAM & 480 & 26.683 & 5.666 & 0.481 & Stella-vSLAM & 507 & 2.996 & 0.345 & 0.226 \\
COLMAP (ERP) & 480 & \third{0.629} & \second{0.013} & \second{0.020} & COLMAP (ERP) & 507 & \third{0.833} & \second{0.122} & \third{0.079} \\
COLMAP (Pinhole) & 480 & \first{0.459} & \first{0.009} & \first{0.018} & COLMAP (Pinhole) & 507 & \second{0.697} & \third{0.122} & \second{0.079} \\
Spherical-SfM (ERP) & 480 & 34.575 & 6.330 & 0.786 & Spherical-SfM (ERP) & 507 & 1.224 & 0.148 & 0.102 \\
\textbf{Ours} & 479 & \second{0.571} & 0.047 & 0.178 & \textbf{Ours} & 506 & 0.976 & \first{0.106} & \first{0.064} \\
\midrule
\multicolumn{5}{>{\columncolor{groupgray}}l}{\textbf{Residence}} & \multicolumn{5}{>{\columncolor{groupgray}}l}{\textbf{Road}} \\
Method & $n$ & ATE$\downarrow$ & RPE-t$\downarrow$ & RPE-r$\downarrow$ & Method & $n$ & ATE$\downarrow$ & RPE-t$\downarrow$ & RPE-r$\downarrow$ \\
\midrule
Scal3R & 844 & 4.364 & 0.197 & 0.414 & Scal3R & 866 & 15.602 & 0.167 & 0.285 \\
360DVO & 844 & 1.583 & \second{0.022} & \second{0.030} & 360DVO & 866 & 6.252 & 0.099 & \third{0.024} \\
Open-vSLAM & 719 & 1.384 & 4.831 & 37.360 & Open-vSLAM & 314 & 21.874 & 8.076 & 47.028 \\
Stella-vSLAM & 495 & 1.113 & 0.170 & 0.194 & Stella-vSLAM & 426 & 11.772 & 1.320 & 1.546 \\
COLMAP (ERP) & -- & -- & -- & -- & COLMAP (ERP) & 866 & \second{2.116} & \first{0.081} & \second{0.022} \\
COLMAP (Pinhole) & 844 & \second{0.193} & \first{0.009} & \first{0.023} & COLMAP (Pinhole) & 866 & \third{2.224} & \second{0.081} & \first{0.020} \\
Spherical-SfM (ERP) & 844 & \third{0.395} & \third{0.044} & 0.043 & Spherical-SfM (ERP) & 866 & \first{2.006} & 0.105 & 0.039 \\
\textbf{Ours} & 843 & \first{0.115} & 0.046 & \third{0.039} & \textbf{Ours} & 865 & 2.585 & \third{0.088} & 0.200 \\
\midrule
\multicolumn{5}{>{\columncolor{groupgray}}l}{\textbf{Residence2}} & \multicolumn{5}{>{\columncolor{groupgray}}l}{\textbf{Mean}} \\
Method & $n$ & ATE$\downarrow$ & RPE-t$\downarrow$ & RPE-r$\downarrow$ & Method & $n$ & ATE$\downarrow$ & RPE-t$\downarrow$ & RPE-r$\downarrow$ \\
\midrule
Scal3R & 944 & 20.089 & 0.336 & 0.338 & Scal3R & 7/7 & 9.794 & 0.282 & 0.340 \\
360DVO & 944 & 1.323 & 0.033 & 0.043 & 360DVO & 7/7 & \third{2.799} & \second{0.104} & \second{0.039} \\
Open-vSLAM & 944 & \second{1.086} & 2.308 & 0.260 & Open-vSLAM & 7/7 & 7.399 & 6.580 & 25.242 \\
Stella-vSLAM & 283 & 1.883 & 0.364 & 0.074 & Stella-vSLAM & 7/7 & 7.424 & 1.329 & 0.470 \\
COLMAP (ERP) & 944 & \third{1.284} & \second{0.031} & \second{0.042} & COLMAP (ERP) & 6/7 & 3.510\,{\scriptsize (6/7)} & 2.575\,{\scriptsize (6/7)} & 1.452\,{\scriptsize (6/7)} \\
COLMAP (Pinhole) & 944 & 1.528 & \first{0.030} & \first{0.042} & COLMAP (Pinhole) & 7/7 & \second{1.736} & \first{0.099} & \first{0.036} \\
Spherical-SfM (ERP) & 944 & 1.691 & \third{0.031} & \third{0.042} & Spherical-SfM (ERP) & 7/7 & 11.779 & 6.915 & 1.418 \\
\textbf{Ours} & 943 & \first{0.708} & 0.048 & 0.045 & \textbf{Ours} & 7/7 & \first{1.697} & \third{0.111} & \third{0.086} \\
\bottomrule
\end{tabular}}
\end{table}

\begin{table}[!ht]

\centering\scriptsize

\setlength\tabcolsep{8pt}

\caption{\textbf{360LOC dense reconstruction by scene.} P/R/F in percent at 20\,cm. Colors rank methods within each scene and metric.}

\label{tab:dense-scenes-360loc}

\resizebox{\linewidth}{!}{

\begin{tabular}{l *{9}{c}}

\toprule

\multirow{2}{*}{Scene} & \multicolumn{3}{c}{Stella-vSLAM-Dense} & \multicolumn{3}{c}{COLMAP (Pinhole)} & \multicolumn{3}{c}{Ours} \\

\cmidrule(lr){2-4}\cmidrule(lr){5-7}\cmidrule(lr){8-10}

 & P$\uparrow$ & R$\uparrow$ & F$\uparrow$ & P$\uparrow$ & R$\uparrow$ & F$\uparrow$ & P$\uparrow$ & R$\uparrow$ & F$\uparrow$ \\

\midrule

Atrium & \third{47.83} & \first{76.16} & \second{58.76} & \first{67.84} & \third{46.63} & \third{55.27} & \second{65.51} & \second{66.92} & \first{66.20} \\

Concourse & \third{77.73} & \first{59.92} & \first{67.67} & \first{85.15} & \second{54.85} & \second{66.72} & \second{83.88} & \third{52.27} & \third{64.40} \\

Hall & \third{71.55} & \first{49.30} & \first{58.38} & \second{77.85} & \third{34.27} & \third{47.59} & \first{85.95} & \second{36.60} & \second{51.34} \\

Piatrium & \third{52.58} & \second{47.53} & \third{49.93} & \first{94.22} & \third{43.40} & \second{59.43} & \second{81.57} & \first{55.71} & \first{66.20} \\

\bottomrule

\end{tabular}}

\end{table}

\begin{table}[!ht]

\centering\scriptsize

\setlength\tabcolsep{8pt}

\caption{\textbf{YUTO MMS dense reconstruction by scene.} P/R/F in percent at 20\,cm. Colors rank methods within each scene and metric.}

\label{tab:dense-scenes-yuto}

\resizebox{\linewidth}{!}{

\begin{tabular}{l *{9}{c}}

\toprule

\multirow{2}{*}{Scene} & \multicolumn{3}{c}{Stella-vSLAM-Dense} & \multicolumn{3}{c}{COLMAP (Pinhole)} & \multicolumn{3}{c}{Ours} \\

\cmidrule(lr){2-4}\cmidrule(lr){5-7}\cmidrule(lr){8-10}

 & P$\uparrow$ & R$\uparrow$ & F$\uparrow$ & P$\uparrow$ & R$\uparrow$ & F$\uparrow$ & P$\uparrow$ & R$\uparrow$ & F$\uparrow$ \\

\midrule

Sequence A & \third{46.13} & \third{39.25} & \third{42.41} & \second{88.09} & \second{49.94} & \second{63.74} & \first{88.77} & \first{62.81} & \first{73.56} \\

Sequence B & \second{18.74} & \second{4.19} & \second{6.85} & \third{10.13} & \third{0.81} & \third{1.50} & \first{48.63} & \first{32.51} & \first{38.97} \\

Sequence C & \second{13.42} & \second{7.10} & \second{9.29} & \third{9.85} & \third{1.36} & \third{2.39} & \first{66.52} & \first{43.20} & \first{52.38} \\

Sequence D & \second{16.84} & \second{7.51} & \second{10.39} & \third{8.51} & \third{0.81} & \third{1.48} & \first{38.31} & \first{24.56} & \first{29.93} \\

\bottomrule

\end{tabular}}

\end{table}

\begin{table}[!ht]

\centering\scriptsize

\setlength\tabcolsep{8pt}

\caption{\textbf{PanoRecon dense reconstruction by scene.} P/R/F in percent at 20\,cm. Colors rank methods within each scene and metric.}

\label{tab:dense-scenes-panorecon}

\resizebox{\linewidth}{!}{

\begin{tabular}{l *{9}{c}}

\toprule

\multirow{2}{*}{Scene} & \multicolumn{3}{c}{Stella-vSLAM-Dense} & \multicolumn{3}{c}{COLMAP (Pinhole)} & \multicolumn{3}{c}{Ours} \\

\cmidrule(lr){2-4}\cmidrule(lr){5-7}\cmidrule(lr){8-10}

 & P$\uparrow$ & R$\uparrow$ & F$\uparrow$ & P$\uparrow$ & R$\uparrow$ & F$\uparrow$ & P$\uparrow$ & R$\uparrow$ & F$\uparrow$ \\

\midrule

Building 1 & \third{22.72} & \third{37.68} & \third{28.34} & \first{86.74} & \first{71.51} & \first{78.39} & \second{70.69} & \second{68.87} & \second{69.77} \\

Building 2 & \third{49.98} & \third{55.64} & \third{52.66} & \first{67.20} & \first{62.37} & \first{64.69} & \second{64.06} & \second{58.56} & \second{61.19} \\

Gym & \third{34.40} & \third{10.87} & \third{16.52} & \second{58.60} & \second{41.62} & \second{48.67} & \first{66.03} & \first{53.86} & \first{59.33} \\

HPC & \third{51.50} & \third{30.84} & \third{38.58} & \first{84.42} & \second{55.84} & \first{67.21} & \second{65.74} & \first{66.36} & \second{66.05} \\

Residence & \third{50.17} & \third{52.52} & \third{51.32} & \first{89.47} & \second{56.71} & \second{69.42} & \second{76.00} & \first{74.44} & \first{75.22} \\

Road & \third{10.89} & \third{10.20} & \third{10.54} & \first{48.96} & \second{34.74} & \first{40.65} & \second{36.14} & \first{40.14} & \second{38.04} \\

Residence2 & \third{36.58} & \third{19.95} & \third{25.82} & \second{63.86} & \second{40.04} & \second{49.22} & \first{70.89} & \first{74.30} & \first{72.56} \\

\bottomrule

\end{tabular}}

\end{table}

\section{Dense reconstruction evaluation}

\label{app:dense-results}
Tables~\ref{tab:dense-scenes-360loc}, \ref{tab:dense-scenes-yuto}, and~\ref{tab:dense-scenes-panorecon} report per-scene dense reconstruction results on 360LOC, YUTO MMS, and PanoRecon, respectively. Additional point-cloud comparisons cover PanoRecon (Figs.~\ref{fig:dense_building}, \ref{fig:dense_residence}, and~\ref{fig:dense_hpc}), 360LOC Concourse (Fig.~\ref{fig:concourse}), and YUTO MMS Sequences B and C (Figs.~\ref{fig:sequenceb} and~\ref{fig:sequencec}).

\end{document}